\documentclass[twoside,11pt]{article}
\usepackage{jmlr2e}
\pdfoutput=1

\ShortHeadings{SID for Evaluating Causal Graphs}{J.~Peters and P.~B\"uhlmann}
\firstpageno{1} 

\usepackage{multirow}
\usepackage{amsfonts}
\usepackage{algorithmic}
\usepackage{algorithm}
\usepackage{amsmath}
\usepackage{amssymb}
\usepackage{comment}
\usepackage{wrapfig}
\usepackage{graphicx}
\usepackage{xcolor,colortbl}
\usepackage{url}
\usepackage{natbib}
\usepackage{tikz}
\usetikzlibrary{arrows}
\usepackage{paralist}
\newdimen\arrowsize
\pgfarrowsdeclare{arcsq}{arcsq}
{
  \arrowsize=0.2pt
  \advance\arrowsize by .5\pgflinewidth
  \pgfarrowsleftextend{-4\arrowsize-.5\pgflinewidth}
  \pgfarrowsrightextend{.5\pgflinewidth}
}
{
  \arrowsize=1.5pt
  \advance\arrowsize by .5\pgflinewidth
  \pgfsetdash{}{0pt} 
  \pgfsetroundjoin   
  \pgfsetroundcap    
  \pgfpathmoveto{\pgfpoint{0\arrowsize}{0\arrowsize}}
  \pgfpatharc{-90}{-140}{4\arrowsize}
  \pgfusepathqstroke
  \pgfpathmoveto{\pgfpointorigin}
  \pgfpatharc{90}{140}{4\arrowsize}
  \pgfusepathqstroke
}

\newcommand{\independent}{\mbox{${}\perp\mkern-11mu\perp{}$}}

\newcommand{\iid}{\overset{\text{iid}}{\sim}}

\DeclareMathOperator*{\SHD}{SHD}
\DeclareMathOperator*{\DNE}{DNE}
\DeclareMathOperator*{\SID}{SID}

\DeclareMathOperator*{\doo}{do}

\newcommand{\C}[1]{\mathcal{#1}}
\newcommand{\B}[1]{\mathbf{#1}}

\newcommand{\la}{{\leftarrow}}

\newcommand{\X}{{\mathbf X}}

\newcommand{\G}{\C{G}}
\newcommand{\HH}{\C{H}}
\newcommand{\CC}{\C{C}}

\newcommand{\law}[1]{\mathcal{L}({#1})}
\newcommand{\lawX}{{\law{\mathbf X}}}
\newcommand{\mean}{{\mathbf E}}

\newcommand{\AN}[2][]{{\B{AN}}^{#1}_{#2}}
\newcommand{\PA}[2][]{{\B{PA}}^{#1}_{#2}}
\newcommand{\pa}[2][]{{\B{pa}}^{#1}_{#2}}
\newcommand{\CH}[2][]{{\B{CH}}^{#1}_{#2}}
\newcommand{\DE}[2][]{{\B{DE}}^{#1}_{#2}}
\newcommand{\ND}[2][]{{\B{ND}}^{#1}_{#2}}

\newcommand{\given}{\,|\,}

\graphicspath{{./}}

\usepackage{color}

\newcommand{\red}{\textcolor{red}}

\begin{document}

\title{Structural Intervention Distance (SID) for Evaluating Causal Graphs}

\author{\name Jonas Peters \email peters@stat.math.ethz.ch\\
\name Peter B\"uhlmann \email buhlmann@stat.math.ethz.ch\\
\addr Seminar for Statistics\\
ETH Zurich\\
Switzerland}

\editor{??}

\maketitle

\begin{abstract} 
Causal inference relies on the structure of a graph, often a directed
acyclic graph (DAG). Different graphs may result in different causal
inference statements and different intervention distributions. To
quantify such differences, we propose a (pre-) distance between DAGs, the
structural intervention distance (SID). The SID is based on a
graphical criterion only and quantifies the closeness
between two DAGs in terms of their corresponding causal inference
statements.
It is therefore
well-suited for evaluating graphs that are used for computing interventions.
Instead of DAGs it is also possible to compare CPDAGs, completed partially directed acyclic graphs that
represent Markov equivalence classes.
Since it differs significantly from the popular
Structural Hamming Distance (SHD), the SID constitutes a valuable additional measure.
We discuss properties of this distance and provide an efficient implementation with software code available on the first author's homepage (an R package is under construction).
\end{abstract}

\section{Introduction}
Given a true causal DAG $\G$, 
we want
to assess the goodness of an estimate $\HH$: more generally, we
want to measure closeness between two DAGs $\G$ and $\HH$. 
The Structural Hamming Distance (SHD, see Definition~\ref{def:shd}) counts the
number of incorrect edges. Although this provides an intuitive distance
between graphs, it does not reflect their capacity for causal
inference. Instead, we propose to count the pairs of vertices $(i,j)$, for
which the estimate $\HH$ correctly predicts intervention
distributions within the class of distributions that are Markov
with respect to $\G$. 
This results in a new (pre-)distance between DAGs, the Structural Intervention Distance, which adds 
valuable additional information to the established SHD.
We are not aware of any directly related idea. 

Throughout this work we consider a finite family of random variables $\X = (X_1, \ldots, X_p)$ with index set $\B{V} := \{1, \ldots, p\}$ (we use capital letters for random variables and bold letters for sets or vectors). We denote their joint distribution by $\lawX$ 
and denote corresponding densities of $\lawX$ with respect to Lebesgue or the counting measure, by
  $p(\cdot)$ (implicitly assuming their existence). We also denote conditional densities and the density of $\law{\B{Z}}$ with $\B{Z} \subset \B{X}$ by $p(\cdot)$.
A graph $\G=(\B{V},\C{E})$ consists of nodes $\B{V}$ and edges $\C{E} \subseteq \B{V}^2$. With a slight abuse of notation we sometimes identify the nodes (or vertices) $j \in \B{V}$ with the variables $X_j$. 
In Appendix~\ref{app:dags}, we provide further terminology regarding directed
acyclic graphs (DAGs) \citep[e.g.][]{Lauritzen1996,Spirtes2000,Koller2009}
which we require in our work.

The rest of this article is organized as follows: Sections~\ref{sec:shd}
and~\ref{sec:do} review the Structural Hamming Distance and the do calculus \citep[e.g.][]{Pearl2009}, respectively. 
In Section~\ref{sec:sid} we introduce the new structural intervention distance,
prove some of its properties and provide possible
extensions. Section~\ref{sec:sim} contains experiments on synthetic data and Section~\ref{sec:imp} describes an efficient implementation of the SID.  

\subsection{Structural Hamming Distance} \label{sec:shd}
The Structural Hamming Distance \citep{Acid2003, Tsamardinos2006} considers
two partially directed acyclic graphs (PDAGs, see appendix)
and counts how many edges do not coincide.
\begin{definition}[Structural Hamming Distance] \label{def:shd}
Let $\mathbb{P}$ be the space of PDAGs over $p$ variables. The Structural
Hamming Distance (SHD) is defined as
\begin{equation*}
\begin{array}{rcll}
\mathrm{SHD}: \; \mathbb{P} \times \mathbb{P} &\rightarrow& \mathbb{N}&\\
(\G,\HH)& \mapsto & \# \{\,(i,j) \in \B{V}^2 \,\given \, \G \text{ and } \HH \text{ do not have the same type}\\
&& \qquad \qquad \qquad \;\; \text{ of edge between } i \text{ and } j\}\,,
\end{array}
\end{equation*}
where edge types are defined in Appendix~\ref{app:dags}.
\end{definition}
Equivalently, we count pairs $(i,j)$, such that 
$\left((i,j) \in \C{E}_{\G} \Delta \C{E}_{\HH} \right)$  or  $\left((j,i) \in \C{E}_{\G} \Delta \C{E}_{\HH}\right)$, where $A \Delta B := (A \setminus B) \cup (B \setminus A)$ is the symmetric difference. Definition~\ref{def:shd} includes a distance between two DAGs since these are special cases of PDAGs. 
In this work, the SHD is primarily used as a measure of reference
when comparing with our new structural intervention distance. A
comparison to other but similar structural distances (e.g. counting only
missing edges) can be found in \citet{deJongh2009}; all distances
they consider are of similar type as SHD. 

\subsection{Intervention Distributions} \label{sec:do}

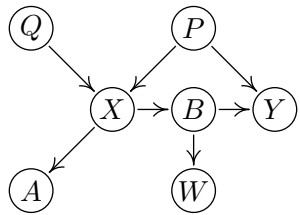
\begin{figure}[ht]
\begin{center}
    \begin{tikzpicture}[scale=1.08, line width=0.5pt, minimum size=0.58cm, inner sep=0.3mm, shorten >=1pt, shorten <=1pt]
    \normalsize
    \draw (0,0) node(x) [circle, draw] {$X$};
    \draw (2,0) node(y) [circle, draw] {$Y$};
    \draw (-1,1) node(p1) [circle, draw] {$Q$};
    \draw (1,1) node(p2) [circle, draw] {$P$};
    \draw (1,0) node(b) [circle, draw] {$B$};
    \draw (-1,-1) node(a) [circle, draw] {$A$};
    \draw (1,-1) node(w) [circle, draw] {$W$};
    \draw[-arcsq] (p1) -- (x);
    \draw[-arcsq] (p2) -- (x);
    \draw[-arcsq] (p2) -- (y);
    \draw[-arcsq] (x) -- (a);
    \draw[-arcsq] (x) -- (b);
    \draw[-arcsq] (b) -- (w);
    \draw[-arcsq] (b) -- (y);
   \end{tikzpicture}
 \end{center}
\caption{The sets $\B{Z}= \{P, Q\}$ and $\B{Z}= \{P, A\}$ are valid adjustment sets for $Y\given \doo(X)$; $\B{Z} = \{P\}$ is
the smallest adjustment set. Any set containing $W$, however, cannot be a valid adjustment set (see Lemma~\ref{lem:adj} below).}
\label{fig:adj}
\end{figure}

Assume that $\lawX$ is absolutely continuous with respect to a product
measure. Then, $\lawX$ is Markov
with respect to $\G$ if and only if the joint density factorizes according
to  
$$ 
p(x_1, \ldots, x_p) = \prod_{j = 1}^p p(x_j \given \B{x}_{\pa[]{j}}),
$$
see for example \citet[][Thm 3.27]{Lauritzen1996}. The intervention
distribution given $\doo(X_i=\hat x_i)$ is then defined as 
$$
p_{\G}(x_1, \ldots, x_p \given \doo(X_i = \hat x_i)) = \prod_{j \neq i} p(x_j \given \B{x}_{\pa[]{j}}) \delta(x_i = \hat x_i) \,.
$$
This, again, is a probability distribution. We can therefore take
expectations or marginalize over some of the variables. One can check (see
proof of Proposition~\ref{prop:main}) that this definition implies\footnote{We sometimes use different letters for the variables in order to avoid subscripts.}
$p_{\G}(y \given \doo(X = \hat x)) = p(y)$ if $Y$ is a parent (or
non-descendant) of $X$; intervening on $X$ does not show any effect on the
distribution of $Y$. If $Y$ is not a parent of $X$, we can compute
(marginalized) intervention distributions by taking into account only a subset of variables from the graph \citep[][Thm 3.2.2]{Pearl2009}. 
\begin{proposition}[Adjustment Formula for Parents] \label{prop:par}
Let $X \neq Y$ be two different nodes in $\G$. If $Y$ is a parent of $X$ then
\begin{equation} \label{eq:par1}
p_{\G}(y \given \doo(X = \hat x)) = p(y) \,.
\end{equation}
If $Y$ is not a parent of $X$ then
\begin{equation} \label{eq:par}
p_{\G}(y \given \doo(X = \hat x)) = \sum_{\pa[]{X}} p(y \given \hat x, \pa[]{X}) \, p(\pa[]{X}) \,.
\end{equation}
\end{proposition}
Whenever we can compute the marginalized intervention distribution $p(y \given \doo(X = \hat x))$ by a summation $\sum_{\B{z}} p(y \given \hat x, \B{z}) \, p(\B{z})$ as in \eqref{eq:par}, we call the set $\B{Z}$ \emph{a valid adjustment set} for the intervention $Y\given \doo(X)$. 
Proposition~\ref{prop:par} states that $\B{Z} = \PA[\G]{X}$ is a valid adjustment set for $Y\given \doo(X)$ (for any $Y$). Figure~\ref{fig:adj} shows that for a given graph there may be other possible adjustment sets.

\section{Structural Intervention Distance} \label{sec:sid}
\subsection{Motivation and Definition}
We propose a new graph-based (pre-)metric, the Structural Intervention
Distance (SID). When comparing graphs (or DAGs in particular), there are
many (pre-)metrics one could consider: an appropriate choice should
depend on the further usage and purpose of the graphs.  
Often one is interested in a causal interpretation of a graph that
enables us to predict the result of interventions.  
We then require a distance that
takes this important goal into account. 
From now on we implicitly assume that
an intervention distribution is computed using adjustment for
parents as in Proposition~\ref{prop:par}; we discuss other choices of
adjustment sets in Section~\ref{sec:altadj}. 
%
The following Example~\ref{ex:shdsid} shows that the SHD
(Definition~\ref{def:shd}) is not well suited for capturing
aspects of the graph that are related to intervention distributions. 
\begin{example} \label{ex:shdsid}
Figure~\ref{fig:ex} shows a true graph $\G$ (left) and two different graphs (e.g. estimates) $\HH_1$ (center) and $\HH_2$ (right). 
\begin{figure}[ht]
  \begin{minipage}[t]{0.3\columnwidth}
   \begin{center}
    \begin{tikzpicture}[scale=1.08, line width=0.5pt, minimum size=0.58cm, inner sep=0.3mm, shorten >=1pt, shorten <=1pt]
    \normalsize
    \draw (-1,0) node(x) [circle, draw] {$X_1$};
    \draw (0,0) node(y) [circle, draw] {$X_2$};
    \draw (1,1) node(z1) [circle, draw] {$Y_1$};
    \draw (1,0) node(z2) [circle, draw] {$Y_2$};
    \draw (1,-1) node(z3) [circle, draw] {$Y_3$};
    \path (x) edge [-arcsq,bend left=30] (z1);
    \path (x) edge [-arcsq,bend left=30] (z2);
    \path (x) edge [-arcsq,bend right=30] (z3);
    \draw[-arcsq] (y) -- (z1);
    \draw[-arcsq] (y) -- (z2);
    \draw[-arcsq] (y) -- (z3);
    \draw[-arcsq] (x) -- (y);
   \end{tikzpicture}\\
true graph $\G$ 
 \end{center}
 \end{minipage}
 \hspace{0.02\columnwidth}
  \begin{minipage}[t]{0.3\columnwidth}
   \begin{center}
    \begin{tikzpicture}[scale=1.08, line width=0.5pt, minimum size=0.58cm, inner sep=0.3mm, shorten >=1pt, shorten <=1pt]
    \normalsize  
    \draw (-1,0) node(x) [circle, draw] {$X_1$};
    \draw (0,0) node(y) [circle, draw] {$X_2$};
    \draw (1,1) node(z1) [circle, draw] {$Y_1$};
    \draw (1,0) node(z2) [circle, draw] {$Y_2$}; 
    \draw (1,-1) node(z3) [circle, draw] {$Y_3$};
    \path (x) edge [-arcsq,bend left=30] (z1);
    \path (x) edge [-arcsq,bend left=30] (z2);
    \path (x) edge [-arcsq,bend right=30] (z3); 
    \draw[-arcsq] (y) -- (z1);
    \draw[-arcsq] (y) -- (z2);
    \draw[-arcsq] (z1) -- (z2); 
    \draw[-arcsq] (y) -- (z3);
    \draw[-arcsq] (x) -- (y);
   \end{tikzpicture}\\
graph $\HH_1$ 
 \end{center}
 \end{minipage}
 \hspace{0.02\columnwidth}
  \begin{minipage}[t]{0.3\columnwidth}
   \begin{center}
    \begin{tikzpicture}[scale=1.08, line width=0.5pt, minimum size=0.58cm, inner sep=0.3mm, shorten >=1pt, shorten <=1pt]
    \normalsize
    \draw (-1,0) node(x) [circle, draw] {$X_1$};
    \draw (0,0) node(y) [circle, draw] {$X_2$};
    \draw (1,1) node(z1) [circle, draw] {$Y_1$};
    \draw (1,0) node(z2) [circle, draw] {$Y_2$};
    \draw (1,-1) node(z3) [circle, draw] {$Y_3$};
    \path (x) edge [-arcsq,bend left=30] (z1);
    \path (x) edge [-arcsq,bend left=30] (z2);
    \path (x) edge [-arcsq,bend right=30] (z3);
    \draw[-arcsq] (y) -- (z1);
    \draw[-arcsq] (y) -- (z2);
    \draw[-arcsq] (y) -- (z3);
    \draw[-arcsq] (y) -- (x);
   \end{tikzpicture}\\
graph $\HH_2$ 
 \end{center}
 \end{minipage}
\caption{Two graphs (center and right) that have the same SHD to the true graph (left), but differ in the SID.}
\label{fig:ex}
\end{figure}
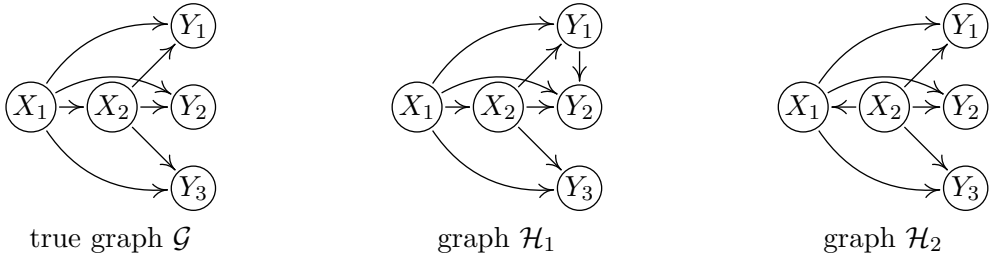
The only difference between $\HH_1$ and $\G$ is the additional edge $Y_1 \rightarrow Y_2$, the only difference between $\HH_2$ and $\G$ is the reversed edge between $X_1$ and $X_2$. The SHD between the true DAG and the others is therefore one in both cases: 
$$
\SHD(\G, \HH_1) = 1 = \SHD(\G, \HH_2)\,.
$$
We now consider a distribution $p(.)$ that is Markov with respect to $\G$ and compute all intervention distributions using parent adjustment~\eqref{eq:par}. 
We will see that these two ``mistakes'' have different impact on the correctness of those intervention distributions.

First, we consider the DAG $\HH_1$. All nodes except for $Y_2$ have the same parent sets in $\G$ and $\HH_1$ and thus, the parent adjustment implies exactly the same formula. Since $X_1$ and $X_2$ are parents of $Y_2$ in both graphs, also the intervention distributions from $Y_2$ to $X_1$ and $X_2$ are correct. We will now argue why $\G$ and $\HH$ agree on the intervention distribution from $Y_2$ to $Y_3$ and from $Y_2$ to $Y_1$. When computing the intervention distribution from $Y_2$ to $Y_3$ in $\HH_1$, we adjust not only for $\{X_1, X_2\}$ as done in $\G$ but also for the additional parent $Y_1$. We thus have to check whether $\{X_1, X_2, Y_1\}$ is a valid adjustment set for $Y_3 \given \doo(Y_2)$. Indeed, since $Y_2 \independent Y_1 \given \{X_1, X_2\}$ (the distribution is Markov with respect to $\G$) we have:
\begin{align*}
p_{\HH_1}(y_3\given \doo(Y_2 = \hat y_2)) &= \sum_{x_1,x_2,y_1} p(y_3\given x_1,x_2,y_1,\hat y_2) p(x_1,x_2,y_1)\\
& = \sum_{x_1,x_2,y_1} \frac{p(x_1,x_2,y_1,\hat y_2,y_3)}{p(\hat y_2 \given x_1,x_2,y_1)}
 = \sum_{x_1,x_2,y_1} \frac{p(x_1,x_2,y_1,\hat y_2,y_3)}{p(\hat y_2 \given x_1,x_2)}\\
& = \sum_{x_1,x_2} p(y_3\given x_1,x_2,\hat y_2) p(x_1,x_2)
 = p_{\G}(y_3\given \doo(Y_2 = \hat y_2))
\end{align*}
It remains to show that $p_{\G}(y_1\given \doo(Y_2 = \hat y_2)) = p(y_1) = p_{\HH_1}(y_1\given \doo(Y_2 = \hat y_2))$, where the last equality is given by~\eqref{eq:par1}.
But since $Y_1 \independent \given X_1, X_2$ it follows from the parent adjustment~\eqref{eq:par} that $p_{\G}(y_1\given \doo(Y_2 = \hat y_2)) = p(y_1)$. 
Thus, all intervention distributions computed in $\HH_1$ agree with those computed in $\G$. Proposition~\ref{prop:sidsuper} shows that this is not a coincidence. It proves that all estimates for which the true DAG is a subgraph correctly predict the intervention distributions.

The ``mistake'' in graph $\HH_2$, namely the reversed edge, is more severe. For computing the correct intervention distribution from $X_2$ to $Y_1$, for example, we need to adjust for the confounder $X_1$, as suggested by the parent adjustment~\eqref{eq:par} applied to $\G$. In $\HH_2$, however, $X_2$ does not have any parent, so there is no variable adjusted for. In general, $\HH_2$ therefore leads to a wrong intervention distribution $p_{\HH_2}(y_1\given \doo(X_2 = \hat x_2)) \neq p_{\G}(y_1\given \doo(X_2 = \hat x_2))$. Also, when computing the intervention distribution from $X_1$ to $Y_i$, $i=1,2,3$, we are adjusting for $X_2$, which is now a parent of $X_1$ in $\HH_2$. Again, this may lead to $p_{\HH_2}(y_i\given \doo(X_1 = \hat x_1)) \neq p_{\G}(y_i\given \doo(X_1 = \hat x_1))$. Further, the intervention distributions from $X_1$ to $X_2$ and from $X_2$ to $X_1$ may not be correct, either. In fact, $\HH_2$ makes eight erroneous predictions for many observational distributions $p(.)$.

The preceding deliberations are reflected by the structural intervention distance we propose below (Definition~\ref{def:sid}). We will see that
$$
\SID(\G, \HH_1) = 0 \neq 8 = \SID(\G, \HH_2)\,.
$$
Furthermore, Proposition~\ref{prop:main} below shows us how to read off the SID from the graph structures.
\end{example}

The following argumentation motivates the formal defintion of the SID.
Given a true DAG $\G$ and an estimate $\HH$, we would like to count the number of intervention distributions, which are computed using the structure of $\HH$, that coincide with the ``true'' intervention distributions inferred from $\G$.
This number, however, depends on the observational distribution over all variables. Since we regard $\G$ as the ground truth we assume that the observational distribution is Markov with respect to $\G$. Consider now a specific distribution that factorizes over all nodes, i.e. all variables are independent (this distribution is certainly Markov with respect to $\G$). Then, $\G$ and $\HH$ agree on all intervention distributions, even though their structure can be arbitrarily different. We therefore consider all distributions that are Markov with respect to $\G$ instead of only one:
we count all pairs of nodes, for which the predicted interventions agree {\it for all} observational distributions that are Markov with respect to $\G$. Those pairs are said to ``correctly estimate'' the intervention distribution.
\begin{definition} \label{def:correct}
Let $\G$ and $\HH$ be DAGs over variables $\B{X} = (X_1, \ldots, X_p)$. For $i \neq j$ we say that the intervention distribution from $i$ to $j$ is \emph{correctly inferred by} $\HH$ \emph{with respect to} $\G$ if
$$
p_{\G}(x_j\given \doo(X_i = \hat x_i)) = p_{\HH}(x_j\given \doo(X_i = \hat x_i)) \quad \forall \lawX \text{ Markov wrt } \G \text{ and } \forall \hat x_i 
$$
Otherwise, that is if
$$
\exists \lawX \text{ Markov wrt } \G \text{ and } \hat x_i \text{ with } \quad p_{\G}(x_j\given \doo(X_i = \hat x_i)) \neq p_{\HH}(x_j\given \doo(X_i = \hat x_i))
$$
we call the intervention distribution from $i$ to $j$ \emph{falsely inferred  by} $\HH$ \emph{with respect to} $\G$. 
Here, $p_{\G}$ and $p_{\HH}$ are computed using parent adjustment as in Proposition~\ref{prop:par} (Section~\ref{sec:altadj} discusses an alternative to parent adjustment).
\end{definition}
%
The SID counts the number of falsely inferred intervention distributions. The definition is independent of any distribution which is crucial to allow for a purely graphical characterization. 
\begin{definition}[Structural Intervention Distance] \label{def:sid}
Let $\mathbb{G}$ be the space of DAGs over $p$ variables. We then define
\begin{equation} \label{eq:SIDdagdag}
\begin{array}{rcl}
\mathrm{SID}: \; \mathbb{G} \times \mathbb{G} &\rightarrow& \mathbb{N}\\
(\G,\HH)& \mapsto &\# \{\,(i,j), i \neq j\;|\;\text{ the intervention distribution from } i \text{ to } j\\
&& \qquad \qquad \qquad \quad \text{ is falsely estimated by } \HH \text{ with respect to } \G \}
\end{array}
\end{equation}
as the \emph{structural intervention distance (SID)}.
\end{definition}
Although the SID is a (pre-)metric, see Section~\ref{sec:propp}, it does not satisfy all properties of a metric, in particular it is not symmetric (see Section~\ref{sec:symm} for a symmetrized version).

\subsection{An Equivalent Formulation}
The SID as defined in~\eqref{eq:SIDdagdag} is difficult to compute. We now provide an equivalent formulation that is based on graphical criteria only.
We will see that for each pair $(i,j)$ the question becomes whether $\PA[\HH]{X_i}$ is a valid adjustment set for the intervention $X_j \, |\, \doo(X_i)$ in graph $\G$. 
\citet{Shpitser2010} prove the following characterization of adjustment
sets.
The reader may think of $\B{Z} = \PA[\G]{X}$, which is always a valid adjustment set, as stated in Proposition~\ref{prop:par}.
\begin{lemma}[Characterization of valid Adjustment Sets] \label{lem:adj}
Consider a DAG $\G = (\B{V},\C{E})$, variables $X,Y \in \B{V}$ and a subset $\B{Z} \subset \B{V}\setminus \{X,Y\}$. Consider the property of $\B{Z}$ w.r.t. $(\G,X,Y)$
\begin{equation*}
(*) \left \{
\begin{array}{c}
\text{In } \G \text{, no } Z \in \B{Z} \text{ is a descendant of any } W \text{ which lies on a directed}\\
\text{path from } X \text{ to } Y \text{ and } \B{Z} \text{ blocks all non-directed paths from } X \text{ to } Y.
\end{array}
\right.
\end{equation*}
We then have the following two statements: 
\begin{itemize}
\item[(i)]
Let $\lawX$ be Markov with respect to $\G$. If $\B{Z}$ satisfies $(*)$ w.r.t. $(\G,X,Y)$, then $\B{Z}$ is a valid adjustment set for $Y \given \doo(X)$.
\item[(ii)]
If $\B{Z}$ does not satisfy $(*)$ w.r.t. $(\G,X,Y)$, then there exists $\lawX$ that is Markov with respect to $\G$ that leads to $p_{\G}(y \given \doo(X = \hat x)) \neq \sum_{\B{z}} p(y \given \hat x, \B{z}) \, p(\B{z})$, meaning $\B{Z}$ is not a valid adjustment set.
\end{itemize}
\end{lemma}
If $Y \not \in \PA[\G]{X}$, then $\B{Z} = \PA[\G]{X}$ satisfies condition $(*)$ and statement $(i)$ reduces to Proposition~\ref{prop:par}. In fact, condition $(*)$ is a slight extension of the backdoor criterion \citep{Pearl2009}. 
It is not surprising that other sets than the parent set work, too. We may adjust for children of $X$, for example, as long as they are not part of a directed path, see Figure~\ref{fig:adj} above. Similarly, we do not have to adjust for parents of $X$ for which all unblocked paths to $Y$ lead through $X$.

Using Lemma~\ref{lem:adj} we obtain the following equivalent definition of
the SID, which is entirely graph-based and will later be exploited for
computation. 
\begin{proposition} \label{prop:main}
The SID has the following equivalent definition.
\begin{equation*}
\SID(\G,\HH) = \# \left\{\,(i,j), i \neq j\,|\, 
\begin{array}{cl}
j \in \DE[\G]{i} & \text{if } j \in \PA[\HH]{i}\\
\PA[\HH]{i} \text{ does not satisfy } (*) \text{ for } (\G,i,j) & \text{if } j \not \in \PA[\HH]{i}
\end{array}
\right\}
\end{equation*}
\end{proposition}
The proof is provided in Appendix~\ref{app:prop:main}; it is based on Lemma~\ref{lem:adj}. 

\subsection{Properties} \label{sec:propp}
We first investigate metric properties of the SID. Let us denote the number of nodes in a graph by $p$ (this is overloading notation but does not lead to any ambiguity). We then have that 
$$
0 \leq \SID(\G, \HH) \leq p\cdot (p-1)
$$ 
and  
$$\G = \HH \Rightarrow \SID(\G, \HH) = 0.$$
The SID therefore satisfies the properties of what is sometimes called a
pre-metric\footnote{A function $d:\mathbb{G} \times \mathbb{G} \rightarrow \mathbb{R}$ is called a premetric if $d(a,b) \geq 0$ and $d(a,a)=0$.}.  

The SID is not symmetric: e.g., for a non-empty graph $\G$ and an empty graph $\HH$, we have that $\SID(\G, \HH) \neq 0 = \SID(\HH, \G)$ (if $\G$ is the empty DAG, all sets of nodes satisfy $(*)$ and are therefore valid adjustment sets).

If $\SID(\G, \HH) = 0$ parent adjustment leads to the same intervention distributions in $\G$ and $\HH$ but it does not necessarily hold that $\G = \HH$. Example~\ref{ex:shdsid} shows graphs $\G \neq \HH_1$ with $\SID(\G, \HH_1) = 0$. 
Using Proposition~\ref{prop:main}, we can characterize the set of DAGs that have structural intervention distance zero to a given true DAG $\G$:
\begin{proposition} \label{prop:sidsuper}
Consider two DAGs $\G$ and $\HH$. We then have 
$$
\SID(\G,\HH) = 0 \quad \Leftrightarrow \quad \G \leq \HH
$$
\end{proposition}
Here, $\G \leq \HH$ means that $\G$ is a subgraph of $\HH$ (see Appendix~\ref{app:dags}).
The proof is provided in Appendix~\ref{app:prop:sidsuper};
it works for any type of adjustment set, not just the parent set (see Section~\ref{sec:altadj}). 
Proposition~\ref{prop:sidsuper} states that $\HH$ can contain many more (additional) edges than $\G$ and still receives an SID of zero. 
Intuitively, the SID counts the number of pairs 
$(i,j)$, such that the intervention distribution inferred from the graph
$\HH$ is wrong; the latter happens if the
estimated set of parents $\PA[\HH]{X_i}$ is not a valid adjustment set in
$\G$. 
If an estimate $\HH$
contains strictly too many edges, i.e. $\G \leq \HH$ and $\pa[\G]{X_i} \subseteq \pa[\HH]{X_i}$ for all $i$, the intervention distributions 
are correct; this follows from
$p(x_j \given x_i, \pa[\HH]{X_i}) = 
p(x_j \given x_i, \pa[\G]{X_i})$, see also Lemma~\ref{lem:adj}.
For computing intervention distributions in practice, we have to
estimate $p(x_j \given x_i, \pa[\HH]{X_i})$ based on
finitely many samples. This can be seen as a
regression task, a well-understood problem in statistics.  
It is therefore a question of the regression or feature selection technique, whether we
see this equality (at least approximately) in practice as well. 
Section~\ref{sec:addedge} shows a simple way to combine the SID with another measure in order to obtain zero distance if and only if the two graphs coincide.\\

The following proposition provides loose and sharp bounds when relating
SID to the SHD: they underline the difference between these two measures. The proof is provided in Appendix~\ref{app:prop:sidshd}.
\begin{proposition}[Relating SID and SHD] \label{prop:sidshd}
Consider two DAGs $\G$ and $\HH$.
\begin{enumerate}
\item[(1a)] When the SHD is zero, the SID is zero, too: 
$$
\SHD(\G, \HH) = 0 \implies \SID(\G, \HH) = 0 
$$
\item[(1b)] We have
$$
\SHD(\G, \HH) = 1 \implies \SID(\G, \HH) \leq 2\cdot(p-1)\,.
$$
This bound is sharp.
\item[(2)] There exists $\G$ and $\HH$ such that $\SID(\G,\HH) = 0$ but
  $\SHD(\G,\HH) = p(p-1)/2$ which achieves the maximal possible
  value. Therefore we cannot bound SHD from SID.  
\end{enumerate}
\end{proposition}

\subsection{Extensions} \label{sec:ext}
\subsubsection{SID between a DAG and a CPDAG} \label{sec:pdag}
Let $\mathbb{C}$ denote the space of CPDAGs (completed partially directed acyclic graphs) over $p$ variables. Some causal inference methods like the PC-algorithm \citep{Spirtes2000} or
Greedy Equivalence Search \citep{Chickering2002} do not output a single
DAG, but rather a completed PDAG~$\CC \in \mathbb{C}$ representing a Markov equivalence
class of DAGs. In order to compute the SID between a (true) DAG $\G$
and an (estimated) PDAG, we can
in principle enumerate all DAGs in the Markov 
equivalence class and compute the SID for each single DAG. This way, we
obtain a vector of distances, instead of a single number, and we can
  compute lower and upper bounds for these distances. 
  
Since the enumeration becomes computationally infeasible with large graph
size, we propose to extend the CPDAG locally. Especially for sparse graphs,
this provides a considerable computational speed-up.  
We make use of the fact that the PDAG $\CC$ represents a Markov equivalence class of DAGs only if each chain
component is chordal \citep{Andersson1997}.   
We extend each chordal chain component $c$ (see Section~\ref{app:dags}) locally to all possible DAGs
$\CC_{c,1}, \ldots, \CC_{c,k}$, leaving the other chain components
undirected \citep{Meek1995}. For each extension $\CC_{c,h}\, (1 \leq h \leq k)$ and for each vertex $i$ within
the chain component $c$, we consider
$$
I(\G,\CC_{c,h})_i := \# \left\{j \neq i\;|\; 
\begin{array}{cl}
X_j \in \DE[\G]{X_i} & \text{if } X_j \in \PA[\CC_{c,h}]{X_i}\\
\PA[\CC_{c,h}]{X_i} \text{ does not satisfy } (*) \text{ for graph } \G & \text{if } X_j \not \in \PA[\CC_{c,h}]{X_i}
\end{array} \right\}\,.
$$
For each chain component $c$, we thus obtain $k$ vectors
$I(\G,\CC_{c,1}), \ldots , I(\G,\CC_{c,k})$ each having $\#c$
entries. We then represent each vector with its sum 
$$
S(\G,\CC_{c,h}) = \sum_{i \in c} I(\G,\CC_{c,h})_i\ \ (h=1,\ldots ,k)
$$
and save the minimum and the maximum over the $k$ values
$$
\min_h S(\G,\CC_{c,h}),\quad \max_h S(\G,\CC_{c,h})\,.
$$
These values correspond to the ``best'' and ``worst'' DAG extensions. We then
report the sum over all minima and the sum over all maxima as lower and
upper bound, respectively 
$$
{\SID}_{\mathrm{lower}}(\G,\CC) = \sum_c \min_h S(\G,\CC_{c,h}),\quad
{\SID}_{\mathrm{upper}}(\G,\CC) = \sum_c \max_h S(\G,\CC_{c,h})\,.
$$
This leads to the extended definition
\begin{equation} \label{eq:SIDdagcpdag}
\begin{array}{rcl}
\mathrm{SID}: \; \mathbb{G} \times \mathbb{C} &\rightarrow& \mathbb{N} \times \mathbb{N}\\
(\G,\CC)& \mapsto & \big({\SID}_{\mathrm{lower}}(\G,\CC), {\SID}_{\mathrm{upper}}(\G,\CC)\big)
\end{array}
\end{equation}
The definition guarantees that the neighborhood orientation of two nodes
do not contradict each other. Both the lower and upper bounds are therefore
met by a DAG member in the equivalence class of $\CC$.

The differences between lower and upper bounds can be quite large. If the true DAG is a (Markov) chain $X_1 \rightarrow \ldots \rightarrow X_p$ of length $p$, the corresponding equivalence class contains the correct DAG resulting in an SID of zero (lower bound); it also includes the reversed chain $X_1 \leftarrow \ldots \leftarrow X_p$ resulting in a maximal SID of $p \cdot (p-1)$.

\begin{figure}
\centering
\includegraphics[width=0.5\textwidth]{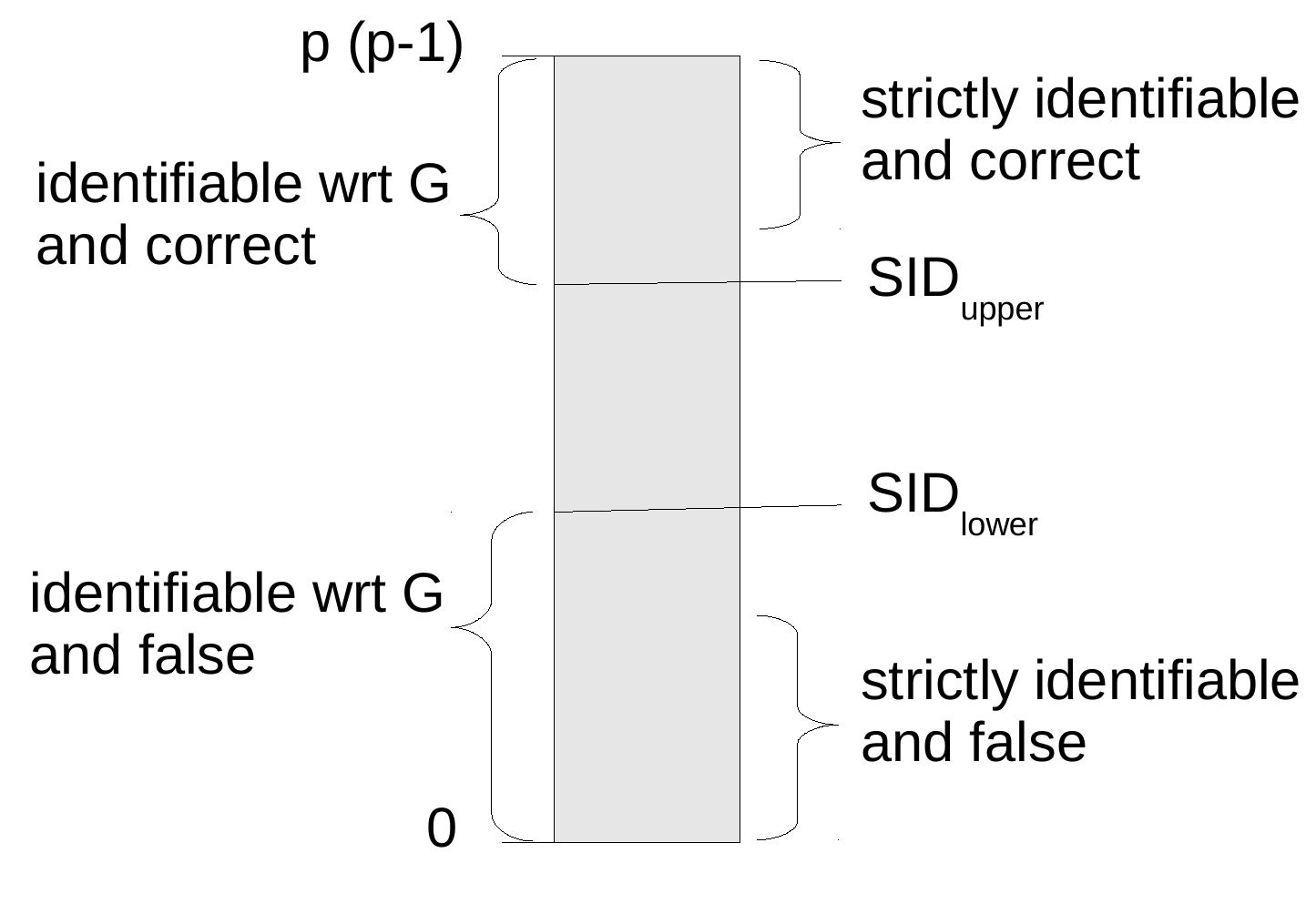}
\caption{This is a visualization of Remark~\ref{rem:cpdag}. It describes the SID between a DAG $\G$ and a CPDAG $\HH$.}
\label{fig:cpdag}
\end{figure}

In order to provide a better intuition for these lower and upper bounds we relate them to ``strictly identifiable'' intervention distributions in the Markov equivalence class.
\begin{definition} \label{def:ident}
Consider a completed partially directed graph $\CC$ and let $\CC_1, \ldots, \CC_k$ be the DAGs contained in the Markov equivalence class represented by $\CC$. We say that the intervention distribution from $i$ to $j$ is 
\begin{itemize}
\item 
\emph{identifiable in $\CC$} if $p_{\CC_g}(x_j \given \doo(X_i = x_i))$ is the same for all $\CC_g \in \{\CC_1, \ldots, \CC_k\}$ and for all distributions $p(.)$ that are Markov with respect to $\CC$.
\item 
\emph{strictly identifiable in $\CC$} if $p_{\CC_g}(x_j \given \doo(X_i = x_i))$ is the same for all $\CC_g \in \{\CC_1, \ldots, \CC_k\}$ and for all distributions $p(.)$.
\item 
\emph{identifiable in $\CC$ w.r.t.  $\G$} if $p_{\CC_g}(x_j \given \doo(X_i = x_i))$ is the same for all $\CC_g \in \{\CC_1, \ldots, \CC_k\}$ and for all distributions $p(.)$ that are Markov w.r.t. $\G$.
\end{itemize}
\end{definition}
Definition~\ref{def:correct} further calls a (strictly) identifiable intervention distribution from $i$ to $j$ estimated correctly if $p_{\G}(x_j \given \doo(X_i = \hat x_i)) = p_{\CC}(x_j \given \doo(X_i = \hat x_i))$ for all $\lawX$ that are Markov with respect to $\G$.
With this notation we have the following remark, which is visualized by Figure~\ref{fig:cpdag}.
\begin{remark} \label{rem:cpdag}
Given a true DAG $\G$ and an estimated CPDAG $\CC$. It then holds (see Figure~\ref{fig:cpdag}) that
\begin{align*}
\# \left\{ \text{interv. distr. that are } 
\begin{array}{c}
\text{ identifiable in }\CC \text{ wrt } \G \text{ and}\\
\text{ inferred falsely by }\CC \text{ wrt } \G
\end{array}
\right\}
&\;= \;{\SID}_{\mathrm{lower}}(\G,\CC)  \\
\# \left\{ \text{interv. distr. that are } 
\begin{array}{c}
\text{ identifiable in }\CC \text{ wrt } \G \text{ and}\\
\text{ inferred correctly by }\CC \text{ wrt } \G
\end{array}
\right\}
&\;= \;p \cdot (p-1) - {\SID}_{\mathrm{upper}}(\G,\CC)\\
\# \left\{ \text{interv. distr. that are } 
\begin{array}{c}
\text{ strictly identifiable in }\CC \text{ and}\\
\text{ inferred falsely by }\CC \text{ wrt } \G
\end{array}
\right\}
&\;\leq \;{\SID}_{\mathrm{lower}}(\G,\CC)  \\
\# \left\{ \text{interv. distr. that are } 
\begin{array}{c}
\text{ strictly identifiable in }\CC \text{ and}\\
\text{ inferred correctly by }\CC \text{ wrt } \G
\end{array}
\right\}
&\;\leq \;p \cdot (p-1) - {\SID}_{\mathrm{upper}}(\G,\CC)\,.
\end{align*}
\end{remark}
Choosing the lower and upper bound to match intervention distributions that are identifiable w.r.t. $\G$ (rather than being strictly identifiable) is a conservative choice. If we us the estimated CPDAGs to provide us with candidate experiments that could reveal nodes with a strong causal effect, we do not want to miss good candidates.

The procedure above fails if $\CC$ is not a {\it completed}
PDAG and therefore does not represent a Markov equivalence class. 
This may happen for some versions of the PC algorithm, when they are based on finitely many data or in the existence of hidden variables.
For each  
node $i$, we can then consider all subsets of undirected neighbors as possible
parent sets and again report lower and upper bounds. The same is done if
the chain component is too large (with more than eight nodes).
These modifications are implemented in our \texttt{R}-code that is available on the first author's homepage.

\subsubsection{SID between a CPDAG and a DAG or CPDAG} \label{sec:pdagpdag}
If we simulate from a linear Gaussian SEM with different error variances, for example, we cannot hope to recover the correct DAG from the joint distribution. If we assume faithfulness, however, it is possible to identify the correct Markov equivalence class. In such situations, one may want to compare the estimated structure with the correct Markov equivalence class (represented by a CPDAG) rather than with the correct DAG. Again, we denote the space of CPDAGs by $\mathbb{C}$. We have defined the $\SID$ on $\mathbb{G} \times \mathbb{G}$ (Definition~\ref{def:sid}) and on $\mathbb{G} \times \mathbb{C}$ (Section~\ref{sec:pdag}). We now want to extend the definition to $\mathbb{C} \times \mathbb{G}$ and $\mathbb{C} \times \mathbb{C}$, where we compare an estimated structure with a true CPDAG $\CC$. The CPDAG $\CC$ represents a Markov equivalence class that includes many different DAGs $\G_1, \ldots, \G_k$. These different DAGs lead to different intervention distributions. The main idea is therefore to consider only those $(i,j)$ for which the intervention distribution from $i$ to $j$ is identifiable in $\CC$ (Definition~\ref{def:ident}).
\citet{MarloesGenBD} introduce a generalized backdoor criterion that can be used to characterize identifiability of intervention distributions. Lemma~\ref{lem:ide} is a direct implication of their Corollary~4.2 and provides a graphical criterion in order to decide whether an intervention distribution is identifiable in a CPDAG. To formulate the result, we define that a path $X_{a_1}, \ldots, X_{a_s}$ in a partially directed graph is \emph{possibly directed} if no edge between $X_{a_f}$ and $X_{a_{f+1}}$, $f \in \{1, \ldots, s-1\}$, is pointing towards $X_{a_f}$.
\begin{lemma} \label{lem:ide}
Let $X_i$ and $X_j$ be two nodes in a CPDAG $\G$. The intervention distribution from $i$  to $j$  is not identifiable if and only if there is a possibly directed path from $X_i$ to $X_j$ starting with an undirected edge.
\end{lemma}
We then define 
\begin{equation} \label{eq:SIDcpdagdag}
\begin{array}{rcl}
\mathrm{SID}: \; \mathbb{C} \times \mathbb{G} &\rightarrow& \mathbb{N}\\
(\CC,\HH)& \mapsto &\# \{\,(i,j), i \neq j\;|\;\text{the interv. distr from $i$ to $j$ is identif. in $\CC$}\\
&& \qquad \qquad \qquad \quad \text{and } \exists \lawX \text{ that is Markov wrt } \CC_1 \in \CC \text{ such that}\\
&& \qquad \qquad \qquad \quad p_{\CC_1}(x_j\given \doo(X_i = \hat x_i)) \neq p_{\HH}(x_j\given \doo(X_i = \hat x_i)) \}
\end{array}
\end{equation}
In a DAG, all effects are identifiable. The definitions then reduce to the case of DAGs~\eqref{eq:SIDdagdag} and~\eqref{eq:SIDdagcpdag}.
The extension to $\mathrm{SID}: \mathbb{C} \times \mathbb{C} \rightarrow \mathbb{N} \times \mathbb{N}$ is completely analogous to~\eqref{eq:SIDdagcpdag} in Section~\ref{sec:pdag} with lower and upper bounds of the SID score~\eqref{eq:SIDcpdagdag} between a true CPDAG and all DAGs in the estimated Markov equivalence class.

\subsubsection{Penalizing additional edges} \label{sec:addedge}
The estimated DAG may have strictly more edges than the true DAG and still receives an SID of zero (Proposition~\ref{prop:sidsuper}). 
We have argued in Section~\ref{sec:propp} that for computing causal inference this fact only introduces statistical problems that can be dealt with if the sample size increases.
In some practical situations, however, it may nevertheless be seen as an unwanted side effect. This problem can be addressed by introducing an additional distance measuring the difference in number of edges between $\G$ and $\HH$. 
$$
\DNE(\G, \HH) = \big| \# \text{edges in } \G - \# \text{edges in } \HH \big|\,.
$$
Here, a directed or undirected edge counts as one edge. For any DAG $\G$ and any DAG~$\HH$, it then follows directly from Proposition~\ref{prop:sidsuper} that
$$
\G = \HH \quad \Leftrightarrow \quad \big(\SID(\G, \HH) = 0 \text{ and } \DNE(\G, \HH) = 0 \big)\,.
$$
Analogously, we have for any DAG $\G$ and any CPDAG~$\CC$
$$
\G \in \CC \quad \Leftrightarrow \quad \big(\SID_{\mathrm{lower}}(\G, \CC) = 0 \text{ and } \DNE(\G, \CC) = 0 \big)\,.
$$


\subsubsection{Symmetrization} \label{sec:symm}
We may also want to compare two DAGs $\G$ and $\HH$, where neither of them can be seen as an estimate of the other.
For these situations we suggest a symmetrized version of the SID:  
$$
\mathrm{SID}_{\text{symm}}(\G, \HH) = \frac{\SID(\G, \HH) + \SID(\HH, \G)}{2}\,.
$$
Although we believe that this version fits most purposes in practice, there are other possibilities to construct symmetric versions of SID. As a slight modification of Definition~\ref{def:sid}, we may also count all pairs $(i,j)$, such that the intervention distributions coincide for all distributions that are Markov with respect to both graphs. 
Note that this would result in a distance that is always zero if one of its arguments is the empty graph, for example.

\subsubsection{Alternative Adjustment Sets} \label{sec:altadj}
\begin{figure}
\begin{center}
\includegraphics[width=0.41\textwidth]{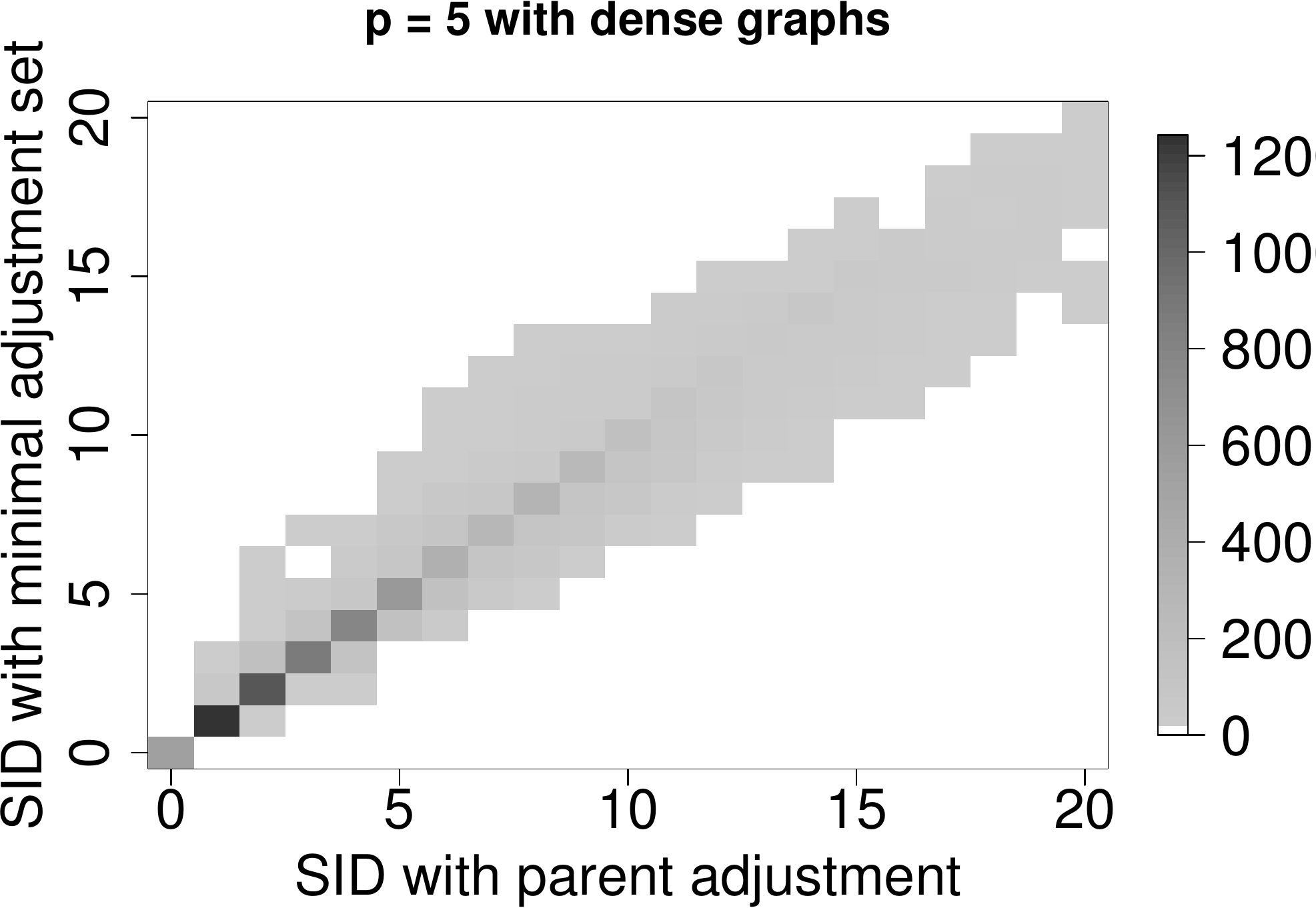}
\end{center}
\caption{The SID between two DAGs is similar when it is computed with parent adjustment or the minimal adjustment set.} 
\label{fig:sidsid2}
\end{figure}

In this work we use the parent set for adjustment. Since it is easy to compute and depends only on the neighbourhood of the intervened nodes it is widely used in practice. 
Any other method to compute adjustment sets in
graphs can be used, too, of course. Choosing an adjustment set of minimal size (see Figure~\ref{fig:adj}) is more difficult to compute but has the advantage of a small conditioning set: \citet{Textor2011} discuss recent advances in efficient computation. In contrast to the parent set, it depends on the whole graph.
Using the experimental setup from Section~\ref{sec:coco} below, we compare the SID computed with parent adjustment with the SID computed with the minimal adjustment set for randomly generated dense graphs of size $p=5$. 
Since the minimal adjustment set need not be unique, we decided to choose the smallest set that is found first by the computational algorithm.
Figure~\ref{fig:sidsid2} shows that the differences between the two values of SID, once computed with parent sets and once computed with minimal adjustment sets, are rather small (especially compared to the differences between SID and SHD, see Section~\ref{sec:coco}). In about $70\%$ of the cases, they are exactly the same. 

\subsubsection{Hidden Variables (future work)} \label{sec:hidden}
If some of the variables are unobserved, not all of the intervention distributions are identifiable from the true DAG. We provide a ``road map'' on how this case can be included in the framework of the SID. As it was done for CPDAGs (Section~\ref{sec:pdagpdag}) we can exclude the non-identifiable pairs from the structural intervention distance. In the presence of hidden variables, the true structure can be represented by an acyclic directed mixed graph (ADMG), for which \citet{Shpitser2006} address the characterization of identifiable intervention distributions. Alternatively, we can regard a maximal ancestral graph (MAG) \citep{Richardson2002} as the ground truth, for which the characterization becomes more difficult. Methods like FCI \citep{Spirtes2000} and its successors \citep{Colombo2012, Claassen2013} output an equivalence class of MAGs that are called partial ancestral graphs (PAGs) \citep{Richardson2002}. To compare an estimated PAG to the true MAG, we would again go through all MAGs represented by the PAG (see Section~\ref{sec:pdag}) and provide lower and upper bounds (as in Section~\ref{sec:pdag}). Future work might show that this can be done efficiently.

\subsubsection{Multiple Interventions (future work)}
The structural intervention distance compares the two graph's predictions
of intervention distributions. Until now, we have only considered
interventions on single nodes. Instead, one may also consider multiple
  interventions. A slightly modified version of
Lemma~\ref{lem:adj} still holds, but the (union of the) parent sets do not
necessarily provide a valid adjustment set, even for the true causal
graph. Instead, one needs to define a ``canonical'' choice of a valid adjustment set.
Furthermore, given a method that computes a valid adjustment set in
the correct graph, one needs to handle the computational complexity that
arises from the large number of possible interventions: for each number $k$
of multiplicity of interventions there are $2^k$ possible intervention sets
and $p-k$ possible target nodes $j$. In total we thus have
$\sum_{k=1}^{p-1} {p \choose k}  (p-k) = p (2^{p-1} - 1)$ intervention
distributions. In practice, one may first address the case of intervening on two nodes, where the number of possible 
intervention distributions is $p(p-1)(p-2)/2$.

\section{Simulations} \label{sec:sim}

\subsection{SID versus SHD} \label{sec:coco}
\begin{figure}[h]
\begin{center}
\includegraphics[width=0.41\textwidth]{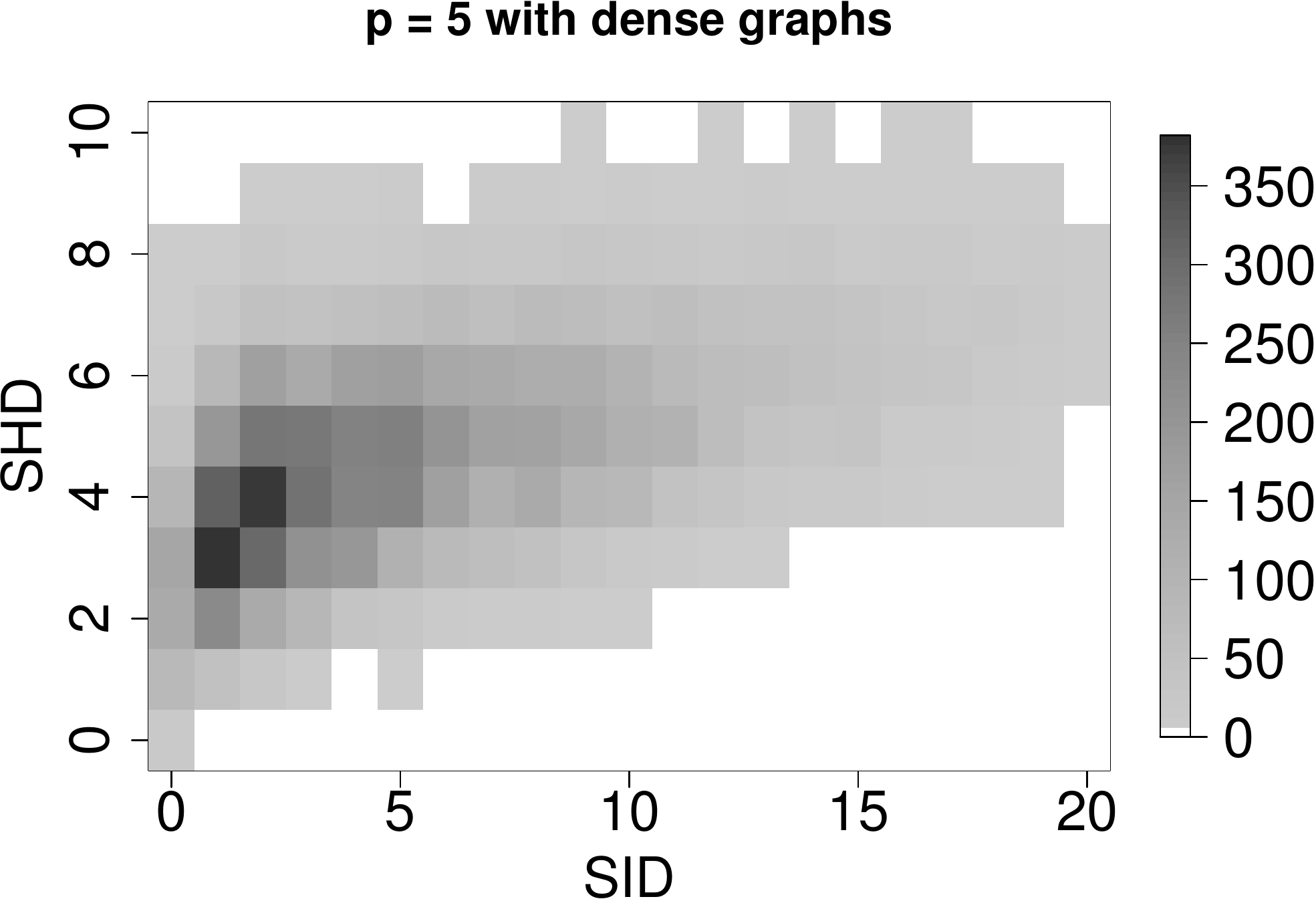}
\hfill
\includegraphics[width=0.41\textwidth]{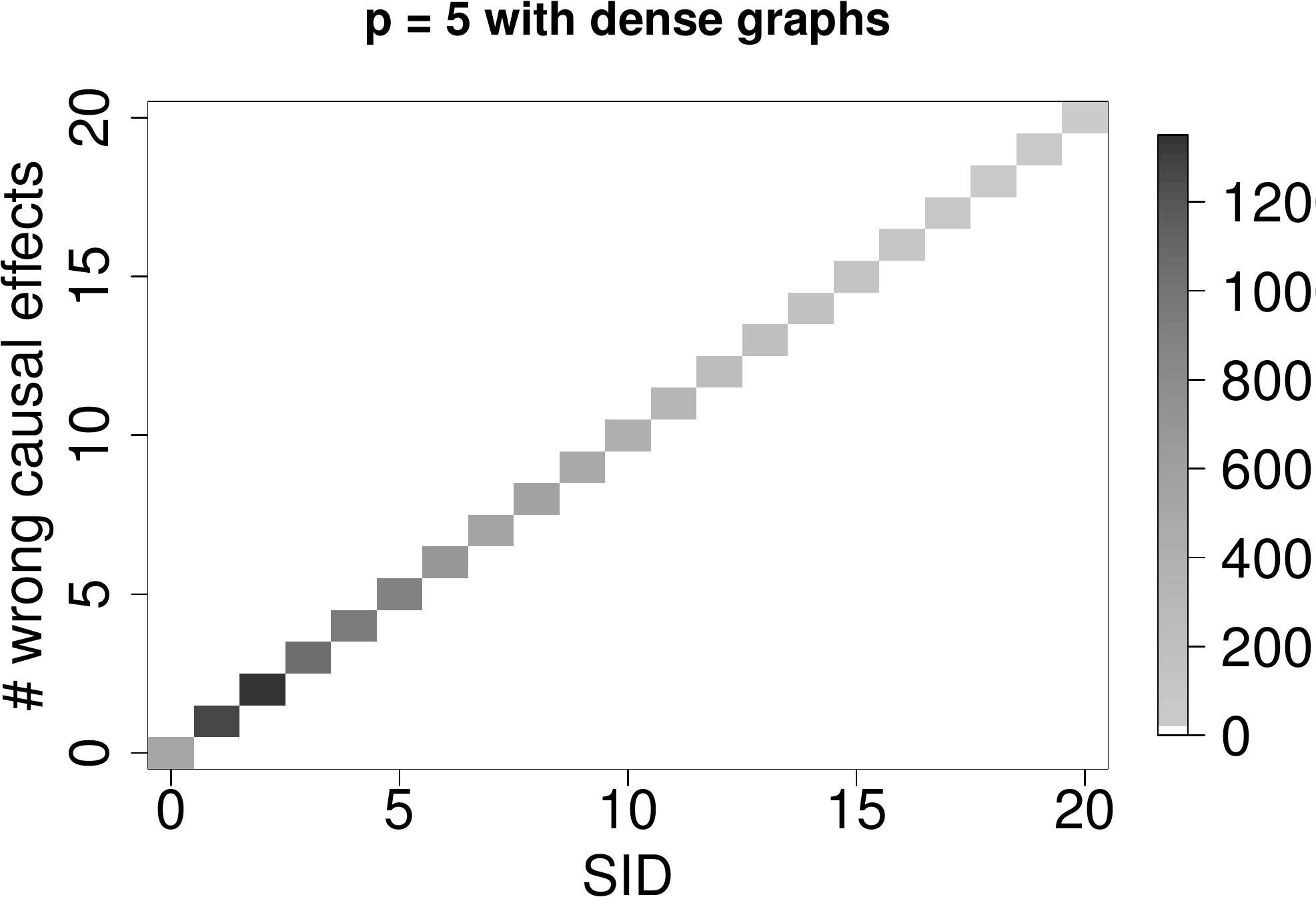}\vspace{0.3cm}\\
\includegraphics[width=0.41\textwidth]{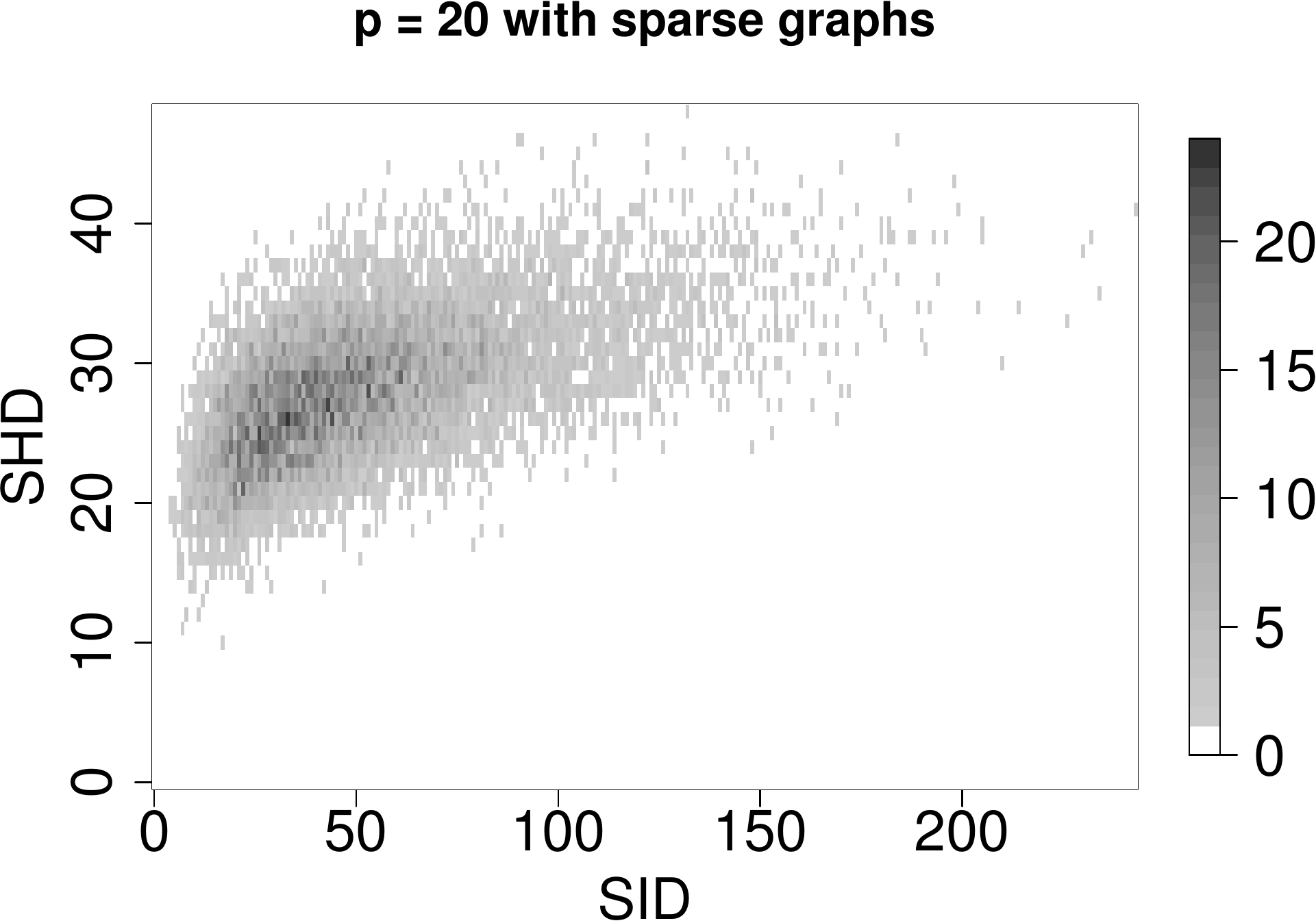}
\hfill
\includegraphics[width=0.41\textwidth]{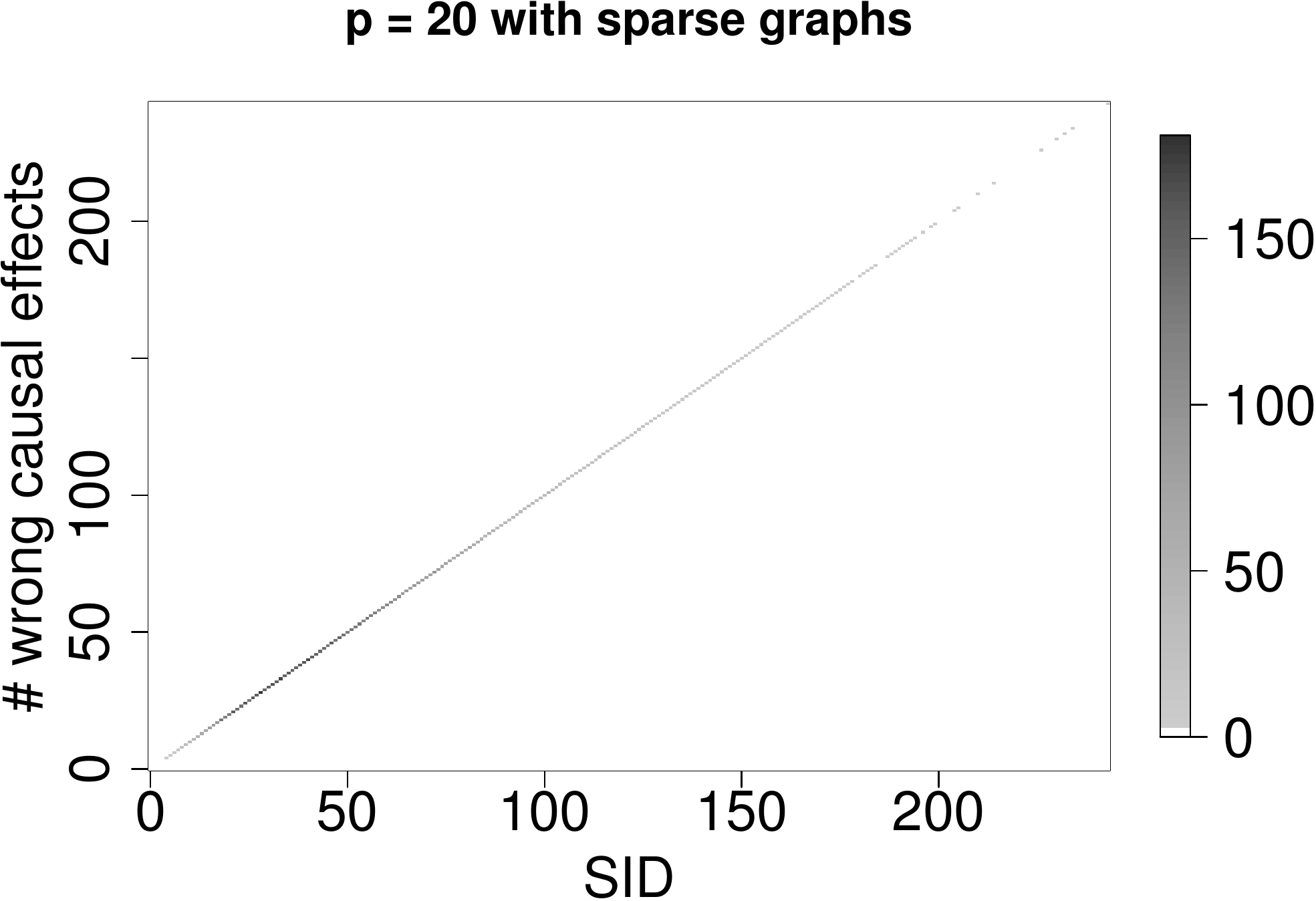}\\
\end{center}
\caption{We generate $10,000$ pairs of random small dense graphs (top) and
  larger sparse graphs (bottom). For each pair of graphs $(\G, \HH)$ we
  also generate a distribution which is Markov w.r.t $\G$. The two-dimensional histograms compare $\SID(\G, \HH)$ with $\SHD(\G, \HH)$ (left) and $\SID(\G, \HH)$ with the number of pairs $(i,j)$, for which the calculated causal effects differ (right). The SID measures exactly the number of wrongly estimated causal effects and thus provides additional and very different information as the SHD.}
\label{fig:sidshd}
\end{figure}

For $p=5$ and for $p=20$ we sample $10,000$ pairs of random DAGs and compute both the SID and the SHD between them.
We consider two probabilities for iid sampling of edges, namely
$p_{\text{connect}} = 1.5/(p-1)$ (resulting in an expected number of
$0.75p$ edges) for a
sparse setting and $p_{\text{connect}} = 0.3$ for a dense
setting. Furthermore, the order of the variables is chosen from a uniformly
distributed permutation among the vertices. 
The left panels in Figure~\ref{fig:sidshd} show two-dimensional histograms with SID and SHD. It is apparent that the SHD and SID constitute very different distance measures.
For example, for SHD equal to a low number such as one or two (see $p=5$ in the dense case), the SID can take on very different values. This indicates, that compared to the SHD, the SID provides additional information that are appropriate for causal inference.
The observations are in par with the bounds provided in Proposition~\ref{prop:sidshd}.\\

For each pair $\G$ and $\HH$ of graphs we also generate a distribution by defining a linear structural equation model  
$$
X_j = \sum_{k \in \pa[\G]{j}} \beta_{jk} X_k + N_j\,, \qquad j= 1, \ldots, p\,,
$$
whose graph is identical to $\G$. We sample the coefficients $\beta_{jk}$
uniformly from $[-1.0;-0.1]$ $\cup$ $[0.1;1.0]$. The noise variables are
normally distributed with mean zero and variance one. Due to the assumption of equal error variances for the error terms, the DAG is identifiable from the distribution \citep{Peters2012}.
With the linear
Gaussian choice we can characterize the true intervention distribution
$p(x_j \given \hat x_i)$ by one number, namely the derivative of the expectation with respect
to $\hat x_i$ (which is also called the total causal
  effect of $X_i$ on $X_j$). Its derivation can be found in Appendix~\ref{app:ce}. We can then compare the intervention distributions from $\G$ and $\HH$ and report the number of pairs $(i,j)$, for which these two numbers differ. For numerical reasons we regard two numbers as different if their absolute difference is larger than $10^{-8}$.
The right panels in Figure~\ref{fig:sidshd} show the comparison to the
SID. In all of the $20,000$ cases, the SID counts exactly the number of
those ``wrong'' causal effects. A priori this is not obvious since
Definition~\ref{def:sid} only requires that there {\it exists} a
distribution that discriminates between the intervention distributions. The
result shown in Figure~\ref{fig:sidshd} suggests that the intervention
distributions differ for {\it most} distributions. Two possible reasons for inequality have indeed small 
probability: (1) a non-detectable difference that is smaller than $10^{-8}$ and (2) vanishing coefficients that would violate faithfulness
\citep[][Thm~3.2]{Spirtes2000}. 
We are not aware of a characterization of the distributions that do not allow to discriminate between the intervention distributions.

\subsection{Comparing Causal Inference Methods}
As in Section~\ref{sec:coco} we simulate sparse random DAGs as ground truth
($100$ times for each value of $p$ and $n$). We again sample $n$ data
points from the corresponding linear Gaussian structural equation model
with equal error variances 
(as above coefficients are uniformly chosen from $[-1;-0.1]$ $\cup$ $[0.1;1]$) and apply
different inference methods. This setting allows us to use the PC algorithm
\citep{Spirtes2000}, conservative PC \citep{Ramsey2006}, greedy equivalent
search (GES) \citep{Chickering2002} and greedy DAG search based on the
assumption of equal error variances ($\text{GDS}_{\text{EEV}}$) \citep{Peters2012}.
Table~\ref{tab:sparse} reports the average SID between the true DAG and the
estimated ones. $\text{GDS}_{\text{EEV}}$ is the only method that outputs a
DAG. All other methods output a Markov equivalence class for which we apply
the extension suggested in Section~\ref{sec:pdag}. 
Additionally, we report the results for a random estimator RAND that does
not take into account any of the data: we sample a DAG as in Section~\ref{sec:coco} but with $p_{\text{connect}}$ uniformly chosen between $0$ and $1$.
Section~\ref{sec:pdag} provides an example, for which the SID can be very different for two
DAGs within the same Markov equivalence class. Table~\ref{tab:sparse} shows that this difference can be quite significant even on average.
While the lower bound often corresponds to a reasonably good estimate, the upper bound may not be better than random guessing for small sample sizes. In fact, for $p=5$ and $n=100$, the distance to the RAND estimate was less than the upper bound for PC in $77$ out of the $100$ experiments (not directly readable from the aggregated numbers in the table). For the SHD, however, the PC algorithm outperforms random guessing; e.g., for $p=5$ and $n=100$, RAND is better than PC in $8$ out of $100$ experiments. 
This supports the idea that the PC algorithm estimates the skeleton of a DAG more reliably than the directions of its edges. 
The results also show how much can be gained when additional assumptions
are appropriate; all methods exploit that the data come from a linear
Gaussian SEM while only $\text{GDS}_{\text{EEV}}$ makes use of the additional constraint of
equal error variances, which leads to identifiability of the DAG from the distribution \citep{Peters2012}. 
We draw different conclusions if we consider the SHD (see Table~\ref{tab:sparseSHD}). 
For $p=40$ and $n=100$, for example, PC performs best with respect to SHD while it is worst with respect to SID.
\begin{table}[h]
\caption{Average SID to true DAG for 100 simulation experiments with standard deviation, for
  different $n$ and $p$. For the methods that output a Markov equivalence
  class (CPC, PC and GES), two rows are shown: they represent DAGs from the
  equivalence class with the smallest and with the largest distance, i.e. the lower and upper bounds in~\eqref{eq:SIDdagcpdag} in Section~\ref{sec:pdag}. Smallest averages are highlighted.} 
\begin{center}
\begin{tabular}{r||c|c|c|c|c} 
\multicolumn{1}{c||}{} & \multicolumn{5}{c}{$n=100$} \\ 
$p$& $\text{GDS}_{\text{EEV}}$ & CPC & PC & GES & RAND \\ \hline
\multirow{2}{*}{$5$}  
&\cellcolor{lightgray} &$2.9 \pm 3.2$&$4.3 \pm 4.7$&$3.3 \pm 4.2$&\multirow{2}{*}{$6.1 \pm 4.0$} \\ 
&  \cellcolor{lightgray}       \multirow{-2}{*}{$1.7 \pm 2.2$}       &$8.8 \pm 5.2$&$7.7 \pm 5.2$&$6.9 \pm 4.6$&                       \\ \hline 
\multirow{2}{*}{$20$}  
&\cellcolor{lightgray} &$22.8 \pm 17.1$&$37.0 \pm 26.8$&$24.4 \pm 17.4$&\multirow{2}{*}{$47.7 \pm 28.8$} \\ 
& \cellcolor{lightgray}  \multirow{-2}{*}{$14.1 \pm 10.5$}                     &$63.3 \pm 38.0$&$52.8 \pm 30.1$&$33.1 \pm 19.1$&                        \\ \hline 
\multirow{2}{*}{$40$}  
&\cellcolor{lightgray} &$56.7 \pm 36.3$&$91.3 \pm 58.3$&$58.9 \pm 34.6$&\multirow{2}{*}{$119.1 \pm 63.8$} \\ 
&\cellcolor{lightgray} \multirow{-2}{*}{$37.2 \pm 27.2$}                       &$147.5 \pm 78.6$&$124.2 \pm 66.4$&$65.9 \pm 36.2$&                         \vspace{0.2cm}   \\
& \multicolumn{5}{c}{$n=1000$} \\ \hline
$p$ & $\text{GDS}_{\text{EEV}}$& CPC & PC & GES& RAND \\ \hline 
\multirow{2}{*}{$5$}  
&\cellcolor{lightgray}  &$1.7 \pm 3.4$&$3.0 \pm 4.7$&$1.9 \pm 3.7$&\multirow{2}{*}{$6.3 \pm 5.0$} \\
& \cellcolor{lightgray}  \multirow{-2}{*}{$0.6 \pm 1.6$}                     &$7.0 \pm 4.8$&$6.7 \pm 4.8$&$6.3 \pm 4.4$& \\ \hline 
\multirow{2}{*}{$20$}  
&\cellcolor{lightgray} &$7.4 \pm 10.3$&$26.4 \pm 28.7$&$8.3 \pm 10.2$&\multirow{2}{*}{$53.1 \pm 36.6$} \\
& \cellcolor{lightgray}  \multirow{-2}{*}{$3.0 \pm 6.7$}                    &$40.0 \pm 28.4$&$40.2 \pm 27.5$&$23.4 \pm 13.1$&\\ \hline 
\multirow{2}{*}{$40$}  
&\cellcolor{lightgray} &$13.8 \pm 12.6$&$62.1 \pm 45.5$&$19.7 \pm 18.7$&\multirow{2}{*}{$132.2 \pm 79.8$}\\ 
& \cellcolor{lightgray}  \multirow{-2}{*}{$7.8 \pm 10.2$}                     &$89.8 \pm 49.5$&$91.9 \pm 49.3$&$43.9 \pm 22.7$&
\end{tabular}
\label{tab:sparse}
\end{center}
\end{table}

\begin{table}[h]
\caption{Same experiment as in Table~\ref{tab:sparse}, this time reporting the average SHD to the true DAG. Smallest averages are highlighted.} 
\begin{center}
\begin{tabular}{r||c|c|c|c|c} 
\multicolumn{1}{c||}{} & \multicolumn{5}{c}{$n=100$}\\ \hline
$p$& $\text{GDS}_{\text{EEV}}$ & CPC  & PC    & GES & RAND  \\ \hline 
$5$&          $1.0 \pm 1.1$    \cellcolor{lightgray}        &$3.1 \pm 1.4$ &$2.6 \pm 1.4$  &$2.7 \pm 1.5$&$6.2 \pm 2.2$    \\  \hline
$20$&         $11.3 \pm 3.1$   \cellcolor{lightgray}        &$13.4 \pm 3.7$&$11.3 \pm 3.1$\cellcolor{lightgray} &$15.0 \pm 3.3$&$96.7 \pm 47.8$         \\ \hline 
$40$&         $43.7 \pm 6.6$           &$27.2 \pm 4.9$&$22.6 \pm 4.6$ \cellcolor{lightgray}&$45.4 \pm 6.1$&$377.9 \pm 195.8$     \vspace{0.2cm}   \\ 
& \multicolumn{5}{c}{$n=1000$}\\ \hline
$p$& $\text{GDS}_{\text{EEV}}$ & CPC  & PC    & GES & RAND  \\ \hline 
$5$&$0.3 \pm 0.6$   \cellcolor{lightgray}      &$2.6 \pm 1.5$ &$2.3 \pm 1.4$ &$2.5 \pm 1.5$ &  $6.0 \pm 2.0$  \\ \hline 
$20$&$2.8 \pm 1.9$     \cellcolor{lightgray}     &$8.6 \pm 2.7$ &$7.7 \pm 2.6$ &$7.8 \pm 2.7$ & $98.4 \pm 50.7$  \\ \hline 
$40$&$10.6 \pm 3.6$    \cellcolor{lightgray}      &$17.0 \pm 3.5$&$15.3 \pm 3.4$&$17.8 \pm 4.0$&$393.5 \pm 189.8$  \\ \hline 

\end{tabular}
\label{tab:sparseSHD}
\end{center}
\end{table}
\subsection{Scalability of the SID} \label{sec:scal}
\begin{figure}[h]
\begin{center}
\includegraphics[width=0.48\textwidth]{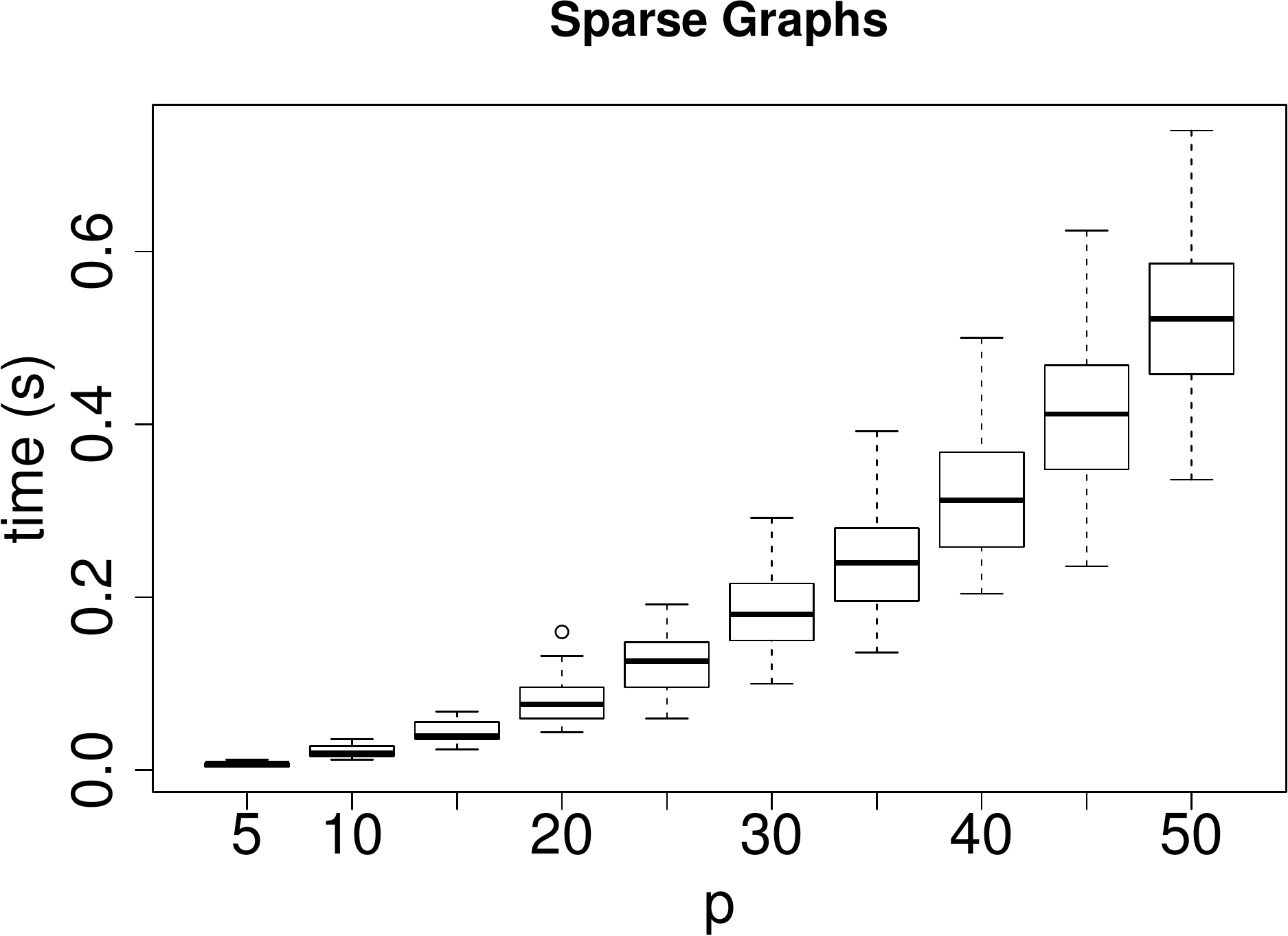}
\hfill
\includegraphics[width=0.48\textwidth]{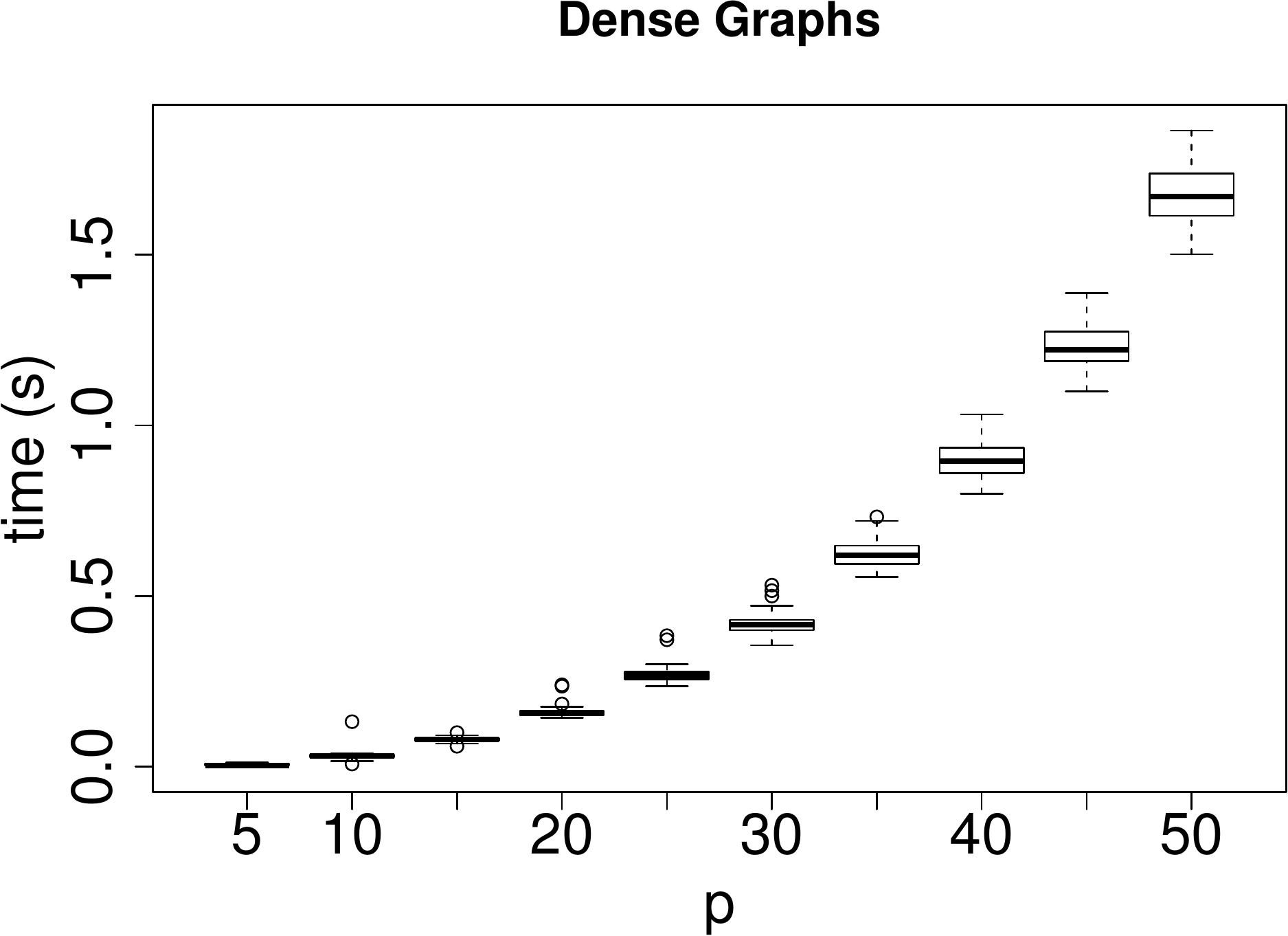}
\end{center}
\caption{Box plots for the processor time needed to compute the SID for one pair of random graphs (averaged over 100
  pairs), for varying $p$ and sparse (left) and dense 
  graphs (right). The computational complexity roughly scales
  quadratic or cubic in $p$ for sparse or dense graphs, respectively.} 
\label{fig:time}
\end{figure}

For different values of $p$ we report here the processor time needed for
computing the SID between two random graphs with $p$ nodes.  
We choose the same setting for sparse and dense graphs as in Section~\ref{sec:coco}.
Figure~\ref{fig:time} shows box plots for $100$ pairs of graphs for each value of $p$ ranging between $5$ and $50$. The figure suggests that the time complexity scales approximately quadratic and cubic in the number of nodes for sparse and dense graphs, respectively\footnote{The experiments were performed on a $64bit$ Ubuntu machine using one core of the Intel Core2 Duo CPU P$8600$ at $2.40$GHz.}.

\section{Implementation} \label{sec:imp}
We sketch here the implementation of the Structural Intervention
Distance while details are presented in Algorithms~\ref{alg:main} and~\ref{alg:adj} in Appendix~\ref{app:alg} using pseudo code. 
The key idea of our algorithm is based on
Proposition~\ref{prop:main}. Condition $(*)$ contains two parts that need
to be checked. Part (1) addressed the issue whether any node from the
conditioning set is a 
descendant of any node on a directed path (see line $28$ in
  Algorithm~\ref{alg:main}). Here, we make use of the $p\times p$
PathMatrix: its entry $(i,j)$ is one if and only if there is a directed
path from $i$ to $j$. This can be computed efficiently by squaring the
matrix $(\mathrm{Id} + \G)$ $\lceil \log_2(p) \rceil$ times since $\G$ is
idempotent; here we denote by $\G$ the adjacency matrix the DAG $\G$. For
part (2) of $(*)$ we check whether the conditioning set 
blocks all non-directed paths from $i$ to $j$ (see line $31$ in
  Algorithm~\ref{alg:main}). It is the
purpose of the function rondp (line $9$ in
  Algorithm~\ref{alg:main}) to compute all nodes that can be
reached on a non-directed path.

Algorithm~\ref{alg:adj}, also presented in the appendix, describes the
function rondp that computes all nodes reachable on
non-directed paths. 
In a breadth-first search we go through all node-orientation combinations
and compute the $2p \times 2p$ reachabilityMatrix. 
Afterwards we compute the corresponding PathMatrix (line~$24$ in
Algorithm~\ref{alg:adj}). We then start with a vector reachableNodes
(consisting of parents and children of node $i$) and read off all reachable
nodes from the reachabilityPathMatrix. We then filter out the nodes that are reachable on a non-directed path. 

Note that in the whole procedure computing the PathMatrix is
computationally the most expensive part. Making sure that this computation
is done only once for all $j$ is one reason why we do not use any
existing implementation (e.g. for $d$-separation).  
The worst case computational complexity for computing the SID between dense matrices is $\mathcal{O}(p \cdot \log_2(p) \cdot f(p))$, where squaring a matrix requires $\mathcal{O}(f(p))$; a naive implementation yields $f(p) = p^3$ while \citet{Coppersmith1987} report $f(p) = \mathcal{O}(p^{2.375477})$, for example.
Sparse matrices lead to improved computational complexities, of course (see also
Section~\ref{sec:scal}).  

We also implemented the steps required for computing the SID between a DAG
and a completed PDAG (both options from Section~\ref{sec:pdag}) using a function
that enumerates all DAGs from partially directed graph. Those steps, however, are not shown in the pseudo code in order to ensure readability. 

Our software code for SID is provided as \texttt{R}-code on the first author's homepage.

\section{Conclusions}
We have proposed a new (pre-) distance, the Structural Intervention
Distance (SID), between directed acyclic graphs and completed partially directed acyclic graphs. 
Since the SID is a one-dimensional measure of distances between high-dimensional objects it does not capture all aspects of the difference.
The SID measures
  ``closeness'' between graphs in terms of their capacities for causal
  effects (intervention distributions). 
It is therefore well suited for evaluating different estimates of causal graphs.
The distance differs significantly
from the widely used Structural Hamming Distance (SHD) and 
can therefore provide a useful complement to existing measures. 
Based on known results for graphical characterization of adjustment sets we have provided a representation of the
SID that enabled us to develop an efficient algorithm for its computation.  
Simulations indicate that in order to draw reliable causal conclusions from
an estimated DAG (i.e. to obtain a small SID), we require more samples than what is suggested by the SHD.  

\section*{Acknowledgments}
We thank Alain Hauser, Preetam Nandy and Marloes Maathuis for helpful discussions. We also thank the anonymous reviewers for their constructive comments.
The research leading to these results has received funding from the People Programme (Marie Curie Actions) of the European Union's Seventh Framework Programme (FP7/2007-2013) under REA grant agreement no $326496$.

\appendix

\section{Terminology for Directed Acyclic Graphs} \label{app:dags}
We summarize here some well known facts about graphs, essentially taken from 
\citep{PetersThesis}.  
Let $\G=(\B{V},\C{E})$ be a graph with $\B{V} := \{1, \ldots, p\}$, $\C{E} \subset \B{V}^2$ and corresponding random variables $\X = (X_1, \ldots, X_p)$.
A graph $\G_1=(\B{V}_1,\C{E}_1)$ is called a {\bf subgraph} of $\G$ if $\B{V}_1 = \B{V}$ and $\C{E}_1 \subseteq \C{E}$; we then write $\G_1 \leq \G$. If additionally, $\C{E}_1 \neq \C{E}$, we call $\G_1$ a {\bf proper subgraph} of $\G$.  
A node $i$ is called a {\bf parent} of $j$ if $(i,j) \in \C{E}$ and a {\bf child} if $(j,i) \in \C{E}$. The set of parents of $j$ is denoted by $\PA[\G]{j}$, the set of its children by $\CH[\G]{j}$. Two nodes $i$ and $j$ are {\bf adjacent} if either $(i,j) \in \C{E}$ or $(j,i) \in \C{E}$.
We call $\G$ {\bf fully connected} if all pairs of nodes are adjacent. We say that there is an undirected edge between two adjacent nodes $i$ and $j$ if $(i,j) \in \C{E}$ and $(j,i) \in \C{E}$; we denote this edge by $i - j$. An edge between two adjacent nodes is directed if it is not undirected; if $(i,j) \in \C{E}$, we denote it by $i \rightarrow j$.
The {\bf skeleton} of $\G$ is the set of all edges without taking the direction into account, that is all $(i,j)$, such that $(i,j) \in \C{E}$ or $(j,i) \in \C{E}$.
The {\bf number of edges} in a graph is the size of the skeleton, i.e. undirected edges count as one.

A {\bf path} $\langle i_1, \ldots, i_n \rangle$ in $\G$ is a sequence of (at least two) distinct vertices
${i_1}, \ldots, {i_n}$, such that there is an edge between ${i_k}$ and ${i_{k+1}}$ for all $k=1, \ldots, n-1$. If $({i_k},{i_{k+1}}) \in \C{E}$ and $({i_{k+1}},{i_{k}}) \notin \C{E}$ for all $k$ we speak of a {\bf directed path} between ${i_1}$ and ${i_n}$ and call ${i_n}$ a {\bf descendant} of ${i_1}$. We denote all descendants of ${i}$ by $\DE[\G]{i}$ and all non-descendants of ${i}$ by $\ND[\G]{i}$.
We call all a node $j$ such that $i$ is a descendant of ${j}$ an {\bf ancestor} of $i$ and denote the set by $\AN[\G]{i}$.
A path $\langle i_1, \ldots, i_n \rangle$ is called a {\bf semi-directed cycle} if $(i_j, i_{j+1}) \in \C{E}$ for $j=1, \ldots, n$ with $i_{n+1}=i_1$ and at least one of the edges is oriented as $i_j \rightarrow i_{j+1}$.
If $({i_{k-1}}, {i_{k}}) \in \C{E}$ and $(i_{k+1}, i_{k}) \in \C{E}$, as well as $({i_{k}}, {i_{k-1}}) \notin \C{E}$ and $(i_{k}, i_{k+1}) \notin \C{E}$, ${i_k}$ is called a {\bf collider} on this path.
$\G$ is called a {\bf partially directed acyclic graph (PDAG)} if there is no directed cycle, i.e. no pair ($j$, $k$), such that there are directed paths from $j$ to $k$ and from $k$ to $j$. 
$\G$ is called a {\bf chain graph} if there is no semi-directed cycle between any pair of nodes. Two nodes $j$ and $k$ in a chain graph are called equivalent if there exists a path between $j$ and $k$ consisting only of undirected edges. A corresponding equivalence class of nodes (i.e. a (maximal) set of nodes that is connected by undirected edges) is called a {\bf chain component}.
$\G$ is called a {\bf directed acyclic graph (DAG)} if it is a PDAG and all edges are directed.
A path in a DAG between ${i_1}$ and ${i_n}$ is {\bf blocked by a set $\B{S}$} (with neither ${i_1}$ nor ${i_n}$ in this set) whenever there is a node ${i_k}$, such that one of the following two possibilities hold:
1. ${i_k} \in \B{S}$ and
${i_{k-1}} \rightarrow {i_k} \rightarrow {i_{k+1}}$ or
${i_{k-1}} \leftarrow {i_k} \leftarrow {i_{k+1}}$ or
${i_{k-1}} \leftarrow {i_k} \rightarrow {i_{k+1}}$; or 2., ${i_{k-1}} \rightarrow {i_k} \leftarrow {i_{k+1}}$ and neither ${i_k}$ nor any of its descendants is in $\B{S}$.
We say that two disjoint subsets of vertices $\B{A}$ and $\B{B}$ are {\bf $d$-separated} by a third (also disjoint) subset $\B{S}$ if every path between nodes in $\B{A}$ and $\B{B}$ is blocked by $\B{S}$.
The joint distribution $\lawX$ is said to be {\bf Markov with respect to the DAG $\G$} if 
$$
\B{A}, \B{B}\; d\text{-sep. by } \B{C} \; \Rightarrow \; {\B{X_A}} \independent {\B{X_B}} \given {\B{X_C}}
$$
for all disjoint sets $\B{A},\B{B},\B{C}$.
$\lawX$ is said to be {\bf faithful to the DAG $\G$} if  
$$
\B{A}, \B{B}\; d\text{-sep. by } \B{C} \; \Leftarrow \; \B{X_A} \independent \B{X_B} \given \B{X_C}
$$
for all disjoint sets $\B{A},\B{B},\B{C}$. Throughout this work, $\independent$ denotes (conditional) independence. 

We denote by $\mathcal{M}(\G)$ the set of distributions that are Markov with respect to $\G$:
$$
\mathcal{M}(\G) := \{\lawX\,:\,\lawX \text{ is Markov wrt }\G \}\,.
$$
Two DAGs $\G_1$ and $\G_2$ are {\bf Markov equivalent} if
$\mathcal{M}(\G_1) = \mathcal{M}(\G_2)$. This is the case if and only if
$\G_1$ and $\G_2$ satisfy the same set of $d$-separations, that means the
Markov condition entails the same set of (conditional) independence
conditions. 
A set of Markov
equivalent DAGs (so-called Markov equivalence 
class) can be represented by a completed PDAG which can be characterized in
terms of a chain graph with undirected and directed edges
\citep{Andersson1997}: this graph has a directed edge if
all members of the Markov equivalence class have such a directed edge, it
has an undirected edge if some members of the Markov equivalence class have
an edge in the same direction and some members have an edge in the other
direction, and it has no edge if all members in the Markov equivalence
class have no corresponding edge. 

\section{Proof of Proposition~\ref{prop:main}} \label{app:prop:main}
Let us denote by $A$ the set of pairs $(i,j)$ appearing in
Definition~\ref{def:sid} and by $B$ the corresponding set of pairs in
Proposition~\ref{prop:main}. We will show that $A=B$.
\begin{itemize}
\item[$ A \subseteq  B$:]
Consider $(i,j) \in A$.

Case (1): If $X_j \in \PA[\HH]{X_i}$, then $p_{\HH}(x_j\given \doo(X_i = \hat x_i)) = p(x_j)$. We will now show that
$p_{\G}(x_j \given \doo(X_i = \hat x_i)) = p(x_j)$ whenever $X_i$ is not an ancestor of $X_j$ in $\G$ (and therefore $X_i$ must be an ancestor of $X_j$). 
\begin{align*}
p_{\G}(x_j \given \doo(X_i = \hat x_i)) &= \int_{\text{anc}(j)} \int_{\text{non-anc}(j)} p(x_1, \ldots, x_p \given \hat x_i) \;d\B{x}_{\text{non-anc}(j)} \;d\B{x}_{\text{anc}(j)}\\
&\overset{(\dagger)}{=} \int_{\text{anc}(j)} \prod_{k \in \text{anc}(j)} p(x_k \given x_{\text{pa}(k)}) \;d\B{x}_{\text{anc}(j)}\\
&= \int_{\text{anc}(j)} \int_{\text{non-anc}(j)} p(x_1, \ldots, x_p) \;d\B{x}_{\text{non-anc}(j)} \;d\B{x}_{\text{anc}(j)}
= p(x_j)
\end{align*}
Equation $(\dagger)$ holds since parents of ancestors of $j$ are ancestors of $j$, too. One can therefore integrate out all non-ancestors (starting at the sink nodes).

Case (2): If, on the other hand, $X_j \not \in \PA[\HH]{X_i}$, then it follows by Lemma~\ref{lem:adj}$(i)$ that $\PA[\HH]{X_i}$ does not satisfy $(*)$. In both cases we have $(i,j) \in B$.
\item[$A \supseteq B$:] Now consider $(i,j) \in B$.

Case (1): If $X_j \in
  \PA[\HH]{X_i}$, then, again, $p_{\HH}(x_j\given \doo(X_i = \hat x_i)) = p(x_j)$ and
  $X_j \in \DE[\G]{X_i}$. Consider a linear Gaussian structural equation
  model with error variances being one and equations $X_k = \sum_{\ell \in \mathbf{pa}_k^{\G}} 1 \cdot X_{\ell} + N_k$, corresponding
  to the graph structure $\G$. It then follows that $p_{\G}(x_j\given \doo(X_i = \hat
  x_i)) \neq p(x_j)$.

Case (2): If $X_j \not \in \PA[\HH]{X_i}$, then $\PA[\HH]{X_i}$ does not satisfy $(*)$ and Lemma~\ref{lem:adj}$(ii)$ implies $p_{\G}(x_j\given \doo( X_i = \hat x_i)) \neq p_{\HH}(x_j\given \doo(X_i = \hat x_i))$.
In both cases we have $(i,j) \in A$.
\end{itemize}

\section{Proof of Proposition~\ref{prop:sidsuper}} \label{app:prop:sidsuper}
\begin{itemize}
\item[$\Leftarrow$:] Assume that $\G \leq \HH$. We will use Proposition~\ref{prop:main} to show that the SID is zero. If $j \in \DE[\G]{i}$ then 
$j \in \DE[\HH]{i}$ which implies that $j \notin \PA[\HH]{i}$. It therefore remains to show that any set $\B{Z}$ that satisfies $(*)$ for $(\HH,i,j)$ satisfies $(*)$ for
for $(\G,i,j)$, too.
The first part of the condition is satisfied since any node that lies on a directed path in $\G$ lies on a directed path in $\HH$. 
The second part holds because any non-directed path in $\G$ is also a path in $\HH$ and must therefore be blocked by $\B{Z}$. If a path is blocked in a DAG it is always blocked in the smaller DAG, too.
\item[$\Rightarrow$:]
Suppose now that $\G$ contains an edge $i \rightarrow j$ and that $i \notin \PA[\HH]{j}$. We now construct an observational distribution $p(.)$ according to $X_k = N_k$ for all $k \neq j$, $X_j = X_i + N_j$ and $N_k \iid \mathcal{N}(0,1)$ for all $k$. This distribution is certainly Markov with respect to $\G$. 
We find for any $\hat x_j$ that $p_{\G}(x_i \given \doo(X_j = \hat x_j)) = p(x_i)$ and at the same time
$p_{\HH}(x_i \given \doo(X_j = \hat x_j)) = p(x_i \given \hat x_j) \neq p(x_i)$. Therefore, the SID is different from zero.
\end{itemize}

\section{Proof of Proposition~\ref{prop:sidshd}} \label{app:prop:sidshd}
The different statements can be proved as follows:
\begin{enumerate}
\item[(1a)] When the SHD is zero, each node has the same set of parents in $\G$ and $\HH$. Therefore all adjustment sets are valid and the SID is zero, too.
\item[(1b)] The bound clearly holds since a SHD of one can change the set of parents of at most two nodes. Extending the example shown in Figure~\ref{fig:ex} from Example~\ref{ex:shdsid} to $p-2$ different $Y$ nodes proves that the bound is sharp.  
\item[(2)] Choosing $\G$ the empty graph and $\HH$ (any) fully connected graph yields the result.
\end{enumerate}

\section{Computing causal effects for linear Gaussian structural
  equation models} \label{app:ce}
Consider a linear Gaussian structural equation model with known
parameters. The covariance matrix $\Sigma_{\X}$ of the $p$ random variables
can then be computed from the structural coefficients and the noise
variances. 
For a given graph we are also able to compute the causal effects analytically. Since the intervention distribution
$\mathcal{L}(X_j \given \doo(X_i = \hat x_i))$ 
is again Gaussian with mean depending linearly on $\hat x_i$ and variance not
depending on $\hat x_i$, we can
summarize it by the so-called {\it causal effect}  
$$
C_{ij} := \frac{\partial}{\partial \hat x} \mean \left[ X_j \given \doo(X_i = \hat x_i) \right]\,.
$$
Let us denote by $\Sigma_{2}$ the submatrix of $\Sigma_{\X}$ with rows and
columns corresponding to $X_i, \PA[]{X_i}$, and by $\Sigma_1$ the
$(1 \times (\# \PA[]{X_i} +1))$-vector corresponding to the row from $X_j$ and columns from $X_i,
\PA[]{X_i}$ of $\Sigma_{\X}$. Then, 
$$
C_{ij} = \Sigma_1 \cdot \Sigma_2^{-1} \cdot (1, 0 \ldots, 0)^T\,.
$$

\newpage
\section{Algorithms} \label{app:alg}
We present here pseudo code of two algorithms for computing the SID. 
\begin{algorithm}[H]
\begin{algorithmic}[1]
\STATE {\bf input} two adjacency matrices $\G$ and $\HH$ of size $p\times p$. \vspace{0.2cm}
\STATE $incorrectCausalEffects \leftarrow$ ZeroMatrix($p,p$)
\STATE PathMatrix $\leftarrow$ computePathMatrix($\G$)
\FOR{$i=1$ {\bfseries to} $p$}
  \STATE $paG \leftarrow$ which($\G[,i]==1$) $\qquad \#$ parents of $i$ in $\G$
  \STATE $paH \leftarrow$ which($\HH[,i]==1$) $\qquad \#$ parents of $i$ in $\HH$
  \STATE $\tilde \G \; \leftarrow \; \G$ without edges leaving $paH$ with a tail (paH $\rightarrow$)  
  \STATE PathMatrix2 $\leftarrow$ computePathMatrix($\tilde \G$)
  \STATE $reachableOnNonDirectedPath \leftarrow$ rondp($\G$,$i$,$paH$,PathMatrix,PathMatrix2)
  \FOR{$j\neq i$ {\bfseries from} $1$ {\bfseries to} $p$} \vspace{0.2cm}
   
\STATE $ijGNull$, $ijHNull$, $finished \leftarrow$ \FALSE
 \IF{PathMatrix$[i,j] == 0$}
   \STATE $ijGNull \leftarrow$ \TRUE $\qquad \#$ $\G$ predicts the causal effect to be zero 
 \ENDIF 
 \IF{$j$ is parent from $i$ in $\HH$}
   \STATE $ijHNull \leftarrow$ \TRUE $\qquad \#$ $\HH$ predicts the causal effect to be zero
 \ENDIF 

 \IF{!$ijGNull$ {\bfseries and} $ijHNull$}
   \STATE incorrectCausalEffects$[i,j]$ $\leftarrow 1$ 
   \STATE $finished \leftarrow$ \TRUE $\qquad \#$ one mistake if only $\HH$ predicts zero
 \ENDIF

 \IF{$ijGNull$ {\bfseries and} $ijHNull$ {\bfseries or} $paG  ==  paH$}
   \STATE $finished \leftarrow$ \TRUE $\qquad \#$ no mistakes if both predictions coincide
 \ENDIF \vspace{0.2cm}

\IF{!finished}
        \STATE $childrenOnDirectedPath \leftarrow$ children of $i$ in $\G$ that have $j$ as a descendant
        \IF{{\bfseries sum}(PathMatrix$[childrenOnDirectedPath,paH]$)$>0$}  
            \STATE incorrectCausalEffects$[i,j]$ $\leftarrow 1$ $\qquad \#$ part (1)
        \ENDIF
                    
        \IF{$reachableOnNonDirectedPath[j]==1$}  
              \STATE incorrectCausalEffects$[i,j]$ $\leftarrow 1$ $\qquad \#$ part (2)
	\ENDIF
\ENDIF \vspace{0.2cm}
\ENDFOR
\ENDFOR \vspace{0.2cm}
\STATE {\bf output} sum(incorrectCausalEffects)
\end{algorithmic}
\caption{Computing structural intervention distance\label{alg:main}}
\end{algorithm}

\begin{algorithm}[H]
\begin{algorithmic}[1]
\STATE {\bf input} adjacency matrix $\G$ of size $p\times p$, node $i$, PaH, PathMatrix, PathMatrix2. \vspace{0.2cm}
\STATE $Pai \leftarrow$ which($\G[,i]==1$) $\qquad \#$ parents of $i$ in $\G$
\STATE $Chi \leftarrow$ which($\G[i,]==1$) $\qquad \#$ children of $i$ in $\G$
\STATE $toCheck \leftarrow Pai+p$ and $Chi$ $\qquad \#$ an index $>p$ indicates that this node is reached\\
\hspace{5.6cm} $\#$ with an outgoing edge, $\leq p$ with an incoming edge\\
\STATE $reachableNodes \leftarrow Pai$ and $Chi$ 
\STATE $reachableOnNonDirectedPath \leftarrow Pai+p\cdot 1_{\text{length}(Pai)}$
\STATE $\G[i,Chi] \leftarrow 0$
\STATE $\G[i,Pai] \leftarrow 0$ \vspace{0.2cm}
\FOR{{\bf all} $currentNode$ {\bf in} $toCheck$}
            \STATE $PacN \leftarrow$ which($\G[,currentNode] == 1$) \vspace{0.1cm}
            
             $ \#$  If one of the Pa of $currentNode$ (cN) is reachable and is not included in $PaH$, \\
             $ \#$  then cN is reachable, too (i.e. $\exists$ path from $i$ that is not blocked by $PaH$).
            \STATE $PacN2 \leftarrow PacN$ setMinus $PaH$
            \STATE reachabilityMatrix$[PacN2,currentNode] \leftarrow 1$ $\qquad \#$ same index rule as for $toCheck$
            \STATE reachabilityMatrix$[PacN2 + p,currentNode] \leftarrow 1$ \vspace{0.1cm}
            
             $ \#$  If $currentNode$ (cN) is reachable with $\rightarrow$ cN and cN is\\ 
             $ \#$  an ancestor of $PaH$, then parents are reachable, too.
            \IF{$currentNode$ is an ancestor of $PaH$}
               \STATE reachabilityMatrix$[currentNode,PacN + p] \leftarrow 1$
               \STATE add $PacN$ to $toCheck$ $\qquad \#$ $toCheck$ is a set; it contains each index only once  
            \ENDIF \vspace{0.1cm}

            $ \#$  If $currentNode$ (cN) is reachable with $\leftarrow$ cN and cN is\\ 
             $ \#$  not in $PaH$, then parents are reachable, too.
            \IF{$currentNode$ is not in $PaH$}
                \STATE reachabilityMatrix$[currentNode + p,PacN + p] \leftarrow 1$    
                \STATE add $PacN$ to $toCheck$ $\qquad \#$ $toCheck$ is a set; it contains each index only once
            \ENDIF \vspace{0.2cm}

            \STATE $\ldots$ $\qquad \#$ Apply analogous rules to the children ChcN of currentNode.
\ENDFOR \vspace{0.2cm}
\STATE reachabilityPathMatrix $\la$ computePathMatrix(reachabilityMatrix)
\STATE update $reachableNodes$ using reachabilityPathMatrix        
\STATE update $reachableOnNonDirectedPath$ using reachabilityPathMatrix \vspace{0.2cm}\\
 $ \#$ We may have missed some nodes: if there is a directed (non-blocked) path \\
 $ \#$ from $i$ to $k$, then all parents of $k$ are reachable from $i$ on a non-directed path.  
\STATE add more nodes to $reachableOnNonDirectedPath$: Use PathMatrix2 to look for nodes $j$ as in $i \rightarrow \ldots \rightarrow k \leftarrow j$ ($k$ being a descendant of $i$ with no node from PaH in between) 
\STATE remove all entries between $k$ and $j$ in computePathMatrix  
\STATE update $reachableOnNonDirectedPath$ using reachabilityPathMatrix 
\STATE $reachableOnNonDirectedPath \leftarrow $ $\{j\,|\, j \text{ or } j+p \in reachableOnNonDirectedPath\}$ \vspace{0.2cm}
\STATE {\bf output} $reachableOnNonDirectedPath$
\end{algorithmic}
\caption{Finding all reachable nodes on non-directed paths (rondp) \label{alg:adj}}
\end{algorithm}

\bibliography{bibliography}

\begin{thebibliography}{21}
\providecommand{\natexlab}[1]{#1}
\providecommand{\url}[1]{\texttt{#1}}
\expandafter\ifx\csname urlstyle\endcsname\relax
  \providecommand{\doi}[1]{doi: #1}\else
  \providecommand{\doi}{doi: \begingroup \urlstyle{rm}\Url}\fi

\bibitem[Acid and de~Campos(2003)]{Acid2003}
S.~Acid and L.~M. de~Campos.
\newblock Searching for {B}ayesian network structures in the space of
  restricted acyclic partially directed graphs.
\newblock \emph{Journal of Artificial Intelligence Research}, 18:\penalty0
  445--490, 2003.

\bibitem[Andersson et~al.(1997)Andersson, Madigan, and Perlman]{Andersson1997}
S.A. Andersson, D.~Madigan, and M.D. Perlman.
\newblock A characterization of {M}arkov equivalence classes for acyclic
  digraphs.
\newblock \emph{Annals of Statistics}, 25:\penalty0 505--541, 1997.

\bibitem[Chickering(2002)]{Chickering2002}
D.M. Chickering.
\newblock Optimal structure identification with greedy search.
\newblock \emph{Journal of Machine Learning Research}, 3:\penalty0 507--554,
  2002.

\bibitem[Claassen et~al.(2013)Claassen, Mooij, and Heskes]{Claassen2013}
T.~Claassen, J.~M. Mooij, and T.~Heskes.
\newblock Learning sparse causal models is not {NP}-hard.
\newblock In \emph{Proceedings of the 29th Annual Conference on {U}ncertainty
  in {A}rtificial {I}ntelligence ({UAI})}, 2013.

\bibitem[Colombo et~al.(2012)Colombo, Maathuis, Kalisch, and
  Richardson]{Colombo2012}
D.~Colombo, M.~Maathuis, M.~Kalisch, and T.~Richardson.
\newblock Learning high-dimensional directed acyclic graphs with latent and
  selection variables.
\newblock \emph{Annals of Statistics}, 40:\penalty0 294--321, 2012.

\bibitem[Coppersmith and Winograd(1987)]{Coppersmith1987}
D.~Coppersmith and S.~Winograd.
\newblock Matrix multiplication via arithmetic progressions.
\newblock In \emph{Proceedings of the 19th annual ACM symposium on Theory of
  computing}, 1987.

\bibitem[de~Jongh and Druzdzel(2009)]{deJongh2009}
M.~de~Jongh and M.~J. Druzdzel.
\newblock A comparison of structural distance measures for causal {B}ayesian
  network models.
\newblock In M.~Klopotek, A.~Przepiorkowski, S.~T. Wierzchon, and
  K.~Trojanowski, editors, \emph{Recent Advances in Intelligent Information
  Systems, Challenging Problems of Science, Computer Science series}, pages
  443--456. Academic Publishing House EXIT, 2009.

\bibitem[Koller and Friedman(2009)]{Koller2009}
D.~Koller and N.~Friedman.
\newblock \emph{Probabilistic Graphical Models: Principles and Techniques}.
\newblock MIT Press, 2009.

\bibitem[Lauritzen(1996)]{Lauritzen1996}
S.~Lauritzen.
\newblock \emph{Graphical Models}.
\newblock Oxford University Press, 1996.

\bibitem[Maathuis and Colombo(2013)]{MarloesGenBD}
M.~Maathuis and D.~Colombo.
\newblock A generalized backdoor criterion.
\newblock \emph{ArXiv e-prints (1307.5636v2)}, 2013.

\bibitem[Meek(1995)]{Meek1995}
C.~Meek.
\newblock Causal inference and causal explanation with background knowledge.
\newblock In \emph{Proceedings of the 11th Annual Conference on {U}ncertainty
  in {A}rtificial {I}ntelligence ({UAI})}, 1995.

\bibitem[Pearl(2009)]{Pearl2009}
J.~Pearl.
\newblock \emph{Causality: Models, Reasoning, and Inference}.
\newblock Cambridge University Press, 2nd edition, 2009.

\bibitem[Peters(2012)]{PetersThesis}
J.~Peters.
\newblock Restricted structural equation models for causal inference.
\newblock PhD Thesis (ETH Zurich), 2012.
\newblock \url{http://dx.doi.org/10.3929/ethz-a-007597940}.

\bibitem[Peters and B{\"u}hlmann(2014)]{Peters2012}
J.~Peters and P.~B{\"u}hlmann.
\newblock Identifiability of {G}aussian structural equation models with equal
  error variances.
\newblock \emph{Biometrika}, 101:\penalty0 219--228, 2014.

\bibitem[Ramsey et~al.(2006)Ramsey, Zhang, and Spirtes]{Ramsey2006}
J.~Ramsey, J.~Zhang, and P.~Spirtes.
\newblock Adjacency-faithfulness and conservative causal inference.
\newblock In \emph{Proceedings of the 22nd Annual Conference on {U}ncertainty
  in {A}rtificial {I}ntelligence ({UAI})}, 2006.

\bibitem[Richardson and Spirtes(2002)]{Richardson2002}
T.~Richardson and P.~Spirtes.
\newblock Ancestral graph {M}arkov models.
\newblock \emph{Annals of Statistics}, 30:\penalty0 962--1030, 2002.

\bibitem[Shpitser and Pearl(2006)]{Shpitser2006}
I.~Shpitser and J.~Pearl.
\newblock Identification of joint interventional distributions in recursive
  semi-markovian causal models.
\newblock In \emph{Proceedings of the 21st National Conference on Artificial
  Intelligence ({AAAI}) - Volume 2}, 2006.

\bibitem[Shpitser et~al.(2010)Shpitser, der Weele, and Robins]{Shpitser2010}
I.~Shpitser, T.~J.~Van der Weele, and J.~M. Robins.
\newblock On the validity of covariate adjustment for estimating causal effects
  (corrected version).
\newblock In \emph{Proceedings of the 26th Annual Conference on {U}ncertainty
  in {A}rtificial {I}ntelligence ({UAI})}, 2010.

\bibitem[Spirtes et~al.(2000)Spirtes, Glymour, and Scheines]{Spirtes2000}
P.~Spirtes, C.~Glymour, and R.~Scheines.
\newblock \emph{Causation, Prediction, and Search}.
\newblock MIT Press, 2nd edition, 2000.

\bibitem[Textor and Liskiewicz(2011)]{Textor2011}
J.~Textor and M.~Liskiewicz.
\newblock Adjustment criteria in causal diagrams: An algorithmic perspective.
\newblock In \emph{Proceedings of the 27th Annual Conference on {U}ncertainty
  in {A}rtificial {I}ntelligence ({UAI})}, 2011.

\bibitem[Tsamardinos et~al.(2006)Tsamardinos, Brown, and
  Aliferis]{Tsamardinos2006}
I.~Tsamardinos, L.~E. Brown, and C.~F. Aliferis.
\newblock The max-min hill-climbing {B}ayesian network structure learning
  algorithm.
\newblock \emph{Machine Learning}, 65:\penalty0 31--78, 2006.

\end{thebibliography}

\end{document}